\documentclass[preprint,12pt]{elsarticle}

\usepackage{amssymb}

\usepackage{graphicx}
\usepackage{subcaption}
\usepackage{multirow}
\usepackage{algorithm}
\usepackage{amsmath,amssymb,amsfonts}
\usepackage{algorithmic}
\usepackage{commath}
\usepackage[normalem]{ulem}
\useunder{\uline}{\ul}{}

\journal{Journal of Computational Science}

\begin{document}

\begin{frontmatter}


\title{Hellinger Distance Weighted Ensemble for Imbalanced Data Stream Classification\tnoteref{t1}}
\tnotetext[t1]{This work was supported by the Polish National Science Center, grant no. 2017/27/B/ST6/01325}

\author[1]{Joanna Grzyb\corref{cor1}}
\ead{joanna.grzyb@pwr.edu.pl}
\author[1]{Jakub Klikowski}
\ead{jakub.klikowski@pwr.edu.pl}
\author[1]{Micha\l~Wo\'zniak}
\ead{michal.wozniak@pwr.edu.pl}

\cortext[cor1]{Corresponding author}
\address[1]{Wroclaw University of Science and Technology, Wybrzeze Wyspianskiego 27, 50-370 Wroclaw, Poland}

\begin{abstract}
The imbalanced data classification remains a vital problem. The key is to find such methods that classify both the minority and majority class correctly. The paper presents the classifier ensemble for classifying binary, non-stationary and imbalanced data streams where the \textit{Hellinger Distance} is used to prune the ensemble. The paper includes an experimental evaluation of the method based on the conducted experiments. The first one checks the impact of the base classifier type on the quality of the classification. In the second experiment, the \textit{Hellinger Distance Weighted Ensemble} (\textsc{hdwe}) method is compared to selected state-of-the-art methods using a statistical test with two base classifiers. The method was profoundly tested based on many imbalanced data streams and obtained results proved the \textsc{hdwe} method's usefulness.
\end{abstract}


\end{frontmatter}



\section{Introduction}
\label{sec:introduction}
Researchers are still working on the imbalanced data stream classification. The problem arises in the reality and there are not many solutions to ensure a high performance. The disproportion among learning instances from different classes can significantly impact classifier learning algorithms \cite{Branco:2016} and usually leads to a bias towards the majority class. The problem of the stationary data imbalance, i.e., when the classes' disproportion is constant, is well established in the literature. However, there are only a few works on the imbalanced data stream where the imbalance ratio may vary over time. 

To illustrate the problem of the imbalanced data stream classification, let us consider the spam filtering test \cite{ditzler2012incremental}. Such a system recognizes which e-mails should be sent to the spam and which ones are appropriate and must be in the recipient's inbox. The number of messages identified as spam can change. At first, a new mailbox user does not get much spam. This is an example of imbalanced data where one class (spam) is less numerous than the other, denoted as the minority class. The second class is the majority class (legitimate e-mails). With time, when the user begins to use the mailbox more often and enters their e-mail address on different web pages, the number of messages increases, both in the first and the second class. It may happen that the relationship between the minority and majority classes, known as the imbalance ratio, increases. For example, the imbalance ratio was 1\%, i.e., 1\% of legitimate e-mails were spam. After some time, this ratio can change 5\%, 10\%, 20\%, etc. Also, it is possible that the amount of spam will increase faster than regular e-mails, and then the minority class will become the majority. In this case, when there are less important messages than spam, which is now the minority class, recognition becomes crucial. The error, in this case, is more expensive. In addition to changing the imbalance between two classes, there may also be a \textit{concept drift} \cite{tsymbal2004problem}. This is a change in the characteristics of e-mails over time, so the data becomes non-stationary. Spammers are constantly trying to change techniques so that more spam goes to the inbox. It means that the classifier's decision boundary changes after a certain period because it needs to learn a new data type and classify it correctly. Combining both of these cases, the imbalance and the \textit{concept drift}, the classification problem becomes very difficult.

The classification of complex data with a single classifier may not be sufficient. However, multiple individual classifiers known as a classifier ensemble can give better classification results. When data has huge volume and arrive continuously it can be called the data stream. It can cause a long time of processing. Data appears continuously, so the system must always be ready to receive it. However, the system cannot store the entire data stream in memory. Therefore only storing the most up-to-date data or data chunks is used. Additionally, we have to implement a forgetting mechanism, which discards old data and thus updates and improves its quality \cite{krawczyk2017ensemble}.

Working on the data type described above, i.e., non-stationary imbalanced data streams, it is possible to refine existing methods that allow statistically significant improvements. It is worth checking how this new method works for the statically and dynamically imbalanced data. When the data is non-stationary, so the \textit{concept drift} may occur, the method should not achieve the poor classification quality. In comparison, the method should work at least as well as the selected state-of-the-art methods. Thus the main contributions of the work are:

\begin{itemize}
    \item The presentation of the new ensemble method \textit{Hellinger Distance Weighted Ensemble} for the imbalanced data stream with the concept drift classification and the computational complexity calculation.
    \item The experimental evaluation of the \textit{Hellinger Distance Weighted Ensemble} and the comparison with the selected state-of-the-art methods.
\end{itemize}

The paper is organized as follows. Section \ref{sec:relatedworks} contains a literature review that provides an overview of imbalanced data stream classification methods. Section \ref{sec:algorithm} describes the new method Hellinger Distance Weighted Ensemble. The experiment plan with the description of data sets and tests as well as the analysis of results and lessons learned can be found in the section \ref{sec:experiments}. The last section \ref{sec:conclusions} concludes the work.

\section{Related works}
\label{sec:relatedworks}

This section discusses the main topics related to the work, starting with an introduction to the imbalanced data analysis task, then an overview of the data stream classification, where inspiring approaches are presented.

\subsection{Imbalanced data}

Data is imbalanced when there is an enormous disproportion between classes. Thus the number of instances in one class is much smaller than in the other. In a two-class classification task, one class is the majority class (negative), and the second one is the minority class (positive) \cite{lopez2013insight}. Particular attention is paid to the classification of the positive instances because the cost of making a mistake can be very high \cite{elkan2001foundations}. 

\subsubsection{Approaches for imbalanced}

According to \cite{Krawczyk:2016}, three types of methods dealing with the class imbalance can be distinguished below.

\begin{itemize}
    \item Algorithm level -- this approach pays special attention to classifying the minority class by adjusting the algorithm in such a way that it is resistant to the skewed distribution \cite{liu2000improving}. It can take into an account the cost of the minority class misclassification. Therefore the algorithm has to consider this \cite{chawla2008automatically}.
    \item Data level -- in this approach, the samples are equalized in both imbalanced classes by data resampling, which allows the use of standard classifiers because the data is already balanced \cite{batista2004study}.
    \item Hybrid approach -- it employs data preprocessing and cost-sensitive learning \cite{krawczyk2014cost}.
\end{itemize}

One of the methods of the first approach is the \textit{Hellinger Distance Decision Tree} (\textsc{hddt}) \cite{cieslak2008learning} \cite{cieslak2012hellinger}. It is the C4.4  decision tree classifier \cite{provost2003tree} employing the \textit{Hellinger Distance} as a splitting criterion. Hence there is no need for additional sampling. Cieslak et al. showed that their method is suitable for the binary classification of imbalanced data sets because it is skew insensitive and robust \cite{cieslak2012hellinger}. However, for balanced data, it is just as good as C4.5 \cite{quinlan2014c4}. Due to the structure of this algorithm, the classification of the high imbalanced data is appropriate.  Noting the positive impact of this approach using the \textit{Hellinger Distance}, the method proposed in this paper focuses on the algorithm level approach. 

The data level approach is focusing on data preprocessing \cite{fernandez2008study}. It is intuitive and usually returns good results. Its main aim is to reduce the number of majority examples (\emph{undersampling}) or to generate new minority instances (\emph{oversampling}). Many of the data sampling methods are independent of classifiers used later \cite{garcia2015data}. However, many researchers use these techniques together with classifiers, e.g., by combining data resampling with the classifier ensemble \cite{chawla2003smoteboost}. Let us concentrate on two main techniques of data sampling:

\begin{itemize}
    \item \textit{Random Oversampling} \cite{wang2009diversity} -- this is a random operation. It involves a duplication of samples in the minority class, which leads to the class balance. Unfortunately, it may lead to overfitting. We may enumerate several modifications of \textit{Random Oversampling} as: \textit{Distributional Random Oversampling} \cite{moreo2016distributional}, \textit{Wrapper-based Random Oversampling} \cite{ghazikhani2012class}, \textit{Generative Oversampling} \cite{liu2007generative}.
    \item \textit{Random Undersampling} \cite{wang2009diversity} -- it removes some of the samples from the majority class randomly to equalize the number of examples from different classes. Because of randomness, it can lead to the elimination of important instances. Then, there could be insufficient samples to learn the model, and the model may be untrained. A few methods can be distinguished: \textit{Inverse Random Undersampling} \cite{tahir2012inverse}, \textit{Tomek Link and Random Undersampling} \cite{elhassan2016classification}, \textit{Clustering-based Undersampling} \cite{lin2017clustering}.
\end{itemize}
    
Based on the classic approach, researchers made some improvements and developed new methods to sample the data. Also, \textit{Synthetic Minority Oversampling Technique} (\textsc{smote}) \cite{chawla2002smote} is oversampling, but unlike random one, \textsc{smote} prevents overfitting by interpolating minority class instances using a \textit{K-Nearest Neighbors} (\textsc{knn}) technique. A weakness of this approach appears when the method's parameters are wrong and instances are added to the minority class. Some better versions of the original \textsc{smote} are: \textit{Borderline \textsc{smote}} \cite{han2005borderline}, \textit{Safe-level-\textsc{smote}} \cite{bunkhumpornpat2009safe}, \textsc{smote-ipf} \cite{saez2015smote}. The \textit{Selective Preprocessing of Imbalanced Data} (\textsc{spider}) \cite{stefanowski2008selective} approach is more complex because it combines oversampling with noisy filtering of majority-class samples depending on the option selected. Wojciechowski and Stefanowski developed a newer version \textsc{spider}3 \cite{wojciechowski2017algorithm}. If there is not minority class, the \textit{Real-value Negative Selection Over-sampling} (\textsc{rnso}) \cite{tao2019real} can be used in the binary classification. It bases on the \textit{Real-value Negative Selection} (\textsc{rns}) algorithm to generate artificial minority instances using feature vectors of majority instances. In the case of the existing few minority instances, \textsc{rns} uses it to initialize detectors. Tao et al. proposed the other over-sampling method \textit{Adaptive Weighted Over-sampling} \cite{tao2020adaptive}. It combines a few approaches: \textit{Density Peaks Clustering}, \textit{Adaptive Sub-cluster Sizing for over-sampling}, \textit{Synthetic Instance Generation} and \textit{Heuristic Filter} to overcome overlapping.

\subsection{Data stream classification}

As we mentioned above, when data comes to the system continuously, it is called the data stream. The data set size is growing very fast, and it could be difficult to analyze this because of a few challenges. Restrictions about time and memory resources are important, especially for real-life problems. Data stream can be divided into small portions of the data called data chunks. This method is known as batch-based or chunk-based learning. Choosing the proper size of the chunk is crucial because it may significantly affect the classification \cite{junsawang2019streaming}.

Instead of chunk-based learning, the algorithm can learn incrementally (online) as well. Training examples arrive one by one at a given time, and they are not kept in memory. The advantage of this solution is the speed of sample processing and the need for smaller memory resources. One of the online methods for imbalanced non-stationary data stream is \textsc{weob} (\textit{Weighted Ensemble of \textsc{oob} and \textsc{uob}}). This method is the ensemble built on the basis of \textsc{oob} (\textit{Oversampling-based Online Bagging}) and \textsc{uob} (\textit{Undersampling-based Online Bagging}). These two basic algorithms counteract the class imbalance in real-time. The \textsc{weob} ensemble contains the best features of oversampling and undersampling. \textsc{uob} works better in static data streams when recognizing the minority class, while \textsc{oob} is more robust for dynamic imbalance changes \cite{wang2014resampling}.

When the data stream is non-stationary and we have limited computational resources, then the well-known cross-validation is insufficient to evaluate the predictive performance \cite{krawczyk2017ensemble}. Two approaches for estimating prediction measures in chunk-based learning can be used instead. \textit{Test-then-train} and \textit{prequential} also apply to metrics described in the section \ref{sec:metrics}.

\begin{itemize}
    \item \textit{Test-then-train} \cite{shaker2015recovery} -- in this technique each data chunk is used first for testing the method and then for training. The model and measurements are updated incrementally after each data chunk.
    \item \textit{Prequential} \cite{krawczyk2017ensemble} -- this is a sequential analysis in which every instance is observed. The model makes a prediction, then the error is estimated. A forgetting mechanism such as a sliding window or fading factors should be implemented to collect selected instances and achieve a more robust estimation.
\end{itemize}

The two following methods were an inspiration to create the \textit{Hellinger Distance Weighted Ensemble} method proposed in this paper.

\subsubsection{Accuracy Weighted Ensembles}
\label{sectionAWE}

Wang et al. conducted their research for the data stream with the \textit{concept drift} and proposed the new \textit{Accuracy Weighted Ensembles} (\textsc{awe}) method \cite{wang2003mining}. They showed that the ensemble classifier outperforms a single classifier. However, for the better data classification, they calculated each classifier's weights in the ensemble and selected the best one. The \textsc{awe} calculates these weights based on the \textit{Accuracy} on the testing data.

The data is divided into equal sized $n$ chunks. The weight of one classifier is expressed by the expected prediction error on the test data, where $DS_n$ is a current $n$th chunk in the data stream in the form $(x, i)$ and $i$ is a true label. Probability that $x$ is an instance of class $i$ given by $k$th classifier is $f^{k}_i(x)$.  The eq. \ref{eq:mse} shows the mean square error (\textsc{mse}) of $k$th classifier.

\begin{equation}
\label{eq:mse}
    MSE_k = \frac{1}{\abs{DS_n}} \sum_{(x, i) \in DS_n} {(1-f^{k}_i(x))^2}
\end{equation}

Also, the MSE of random classification is needed for the weight of the classifier. In the two-class problem $MSE_r=0.25$. The final weight for $k$th classifier is shown in the eq. \ref{eq:wk}.

\begin{equation}
\label{eq:wk}
    w_k = MSE_r - MSE_k
\end{equation}

After building the ensemble, the worst classifiers are removed, ensuring that the classifiers' number is never greater than the level specified when the algorithm was called in the ensemble. Such classifiers are the best possible so that the quality of the classification can be improved. This idea of \textsc{awe} method was used in this work to build \textsc{hdwe}.
 
\subsubsection{Hellinger Distance}
\label{sectionHD}

The \textit{Hellinger Distance} (\textsc{hd}) measures distributional divergence using the Bhattacharyya coefficient. In other words, it is a similarity between the probabilities $P_1$ and $P_2$. This measure is used as a decision tree splitting criterion as \textsc{hddt} method \cite{cieslak2008learning}. Cieslak et al. have proved that the \textit{Hellinger Distance} is skew insensitive. In their next article \cite{cieslak2012hellinger}, they extended \textsc{hddt} research and explored the advantage of their algorithms using isometric lines. The analysis of the obtained results led them to conclude that the increasing imbalance rate between classes has no impact on the algorithm's quality. The \textit{Hellinger Distance} using the \textit{True Positive Rate} (\textsc{tpr}) and the \textit{False Positive Rate} (\textsc{fpr}) presented in the section \ref{sec:metrics} is formulated by Cieslak et al. as follows:

\begin{equation}
\label{eq:HD}
d_H(TPR, FPR) = \sqrt{(\sqrt{TPR}-\sqrt{FPR})^2 + (\sqrt{1-TPR}-\sqrt{1-FPR})^2}
\end{equation}

The proposed \textsc{hdwe} method uses it to prune the committee.

\subsection{Imbalanced data stream}

It has been proven that the use of many individual classifiers assigned different weights can improve the classification's performance. This is called the classifier ensemble or the committee of classifiers \cite{polikar2006ensemble}. Weights for individual classifiers in the ensemble can be set based on various factors, e.g., performance weighting \cite{rokach2010ensemble}. The ensemble has many applications and one of them is the classification of imbalanced data sets \cite{liu2008exploratory}. Based on \cite{galar2011review}, in which the authors presented an overview along with the taxonomy for ensemble methods for the imbalanced problem, selected classifiers are as follows:

\begin{itemize}
    \item Cost-Sensitive Boosting Ensembles
    \begin{itemize}
        \item \textit{AdaCost} \cite{fan1999adacost}
        \item \textit{\textsc{csb}1}, \textit{\textsc{csb}2} \cite{ting2000comparative}
        \item \textit{RareBoost} \cite{joshi2001evaluating}
        \item \textit{AdaC1}, \textit{AdaC2}, \textit{AdaC3} \cite{sun2007cost}
        \item \textit{Self-adaptive cost weights-based SVM cost-sensitive ensemble} \cite{tao2019self}
    \end{itemize}
    \item Boosting Ensemble Learning
    \begin{itemize}
        \item \textit{\textsc{smoteb}oost} \cite{chawla2003smoteboost}
        \item \textit{\textsc{msmoteb}oost} \cite{hu2009msmote}
        \item \textit{\textsc{rusb}oost} \cite{seiffert2009rusboost}
        \item \textit{DataBoost-\textsc{im}} \cite{guo2004learning}
    \end{itemize}
    \item Bagging Ensemble Learning
    \begin{itemize}
        \item \textit{\textsc{smoteb}agging} \cite{wang2009diversity}
        \item \textit{QuasiBagging} \cite{chang2003statistical}
        \item \textit{Asymetric Bagging} \cite{tao2006asymmetric}
        \item \textit{Roughly Balanced Bagging} \cite{hido2009roughly}
        \item \textit{Partitioning} \cite{chan1998learning}
        \item \textit{Bagging Ensemble Variation} \cite{li2007classifying}
        \item \textit{IIVotes} \cite{blaszczynski2010integrating}
        \item \textit{Stratified Bagging and dynamic ensemble selection method} \cite{zyblewski2020preprocessed}
    \end{itemize}
    \item Hybrid Ensemble Learning
    \begin{itemize}
        \item \textit{EasyEnsemble} \cite{liu2008exploratory}
        \item \textit{BalanceCascade} \cite{liu2008exploratory}
    \end{itemize}
\end{itemize}

When the data set is additionally non-stationary, classifying instances into the appropriate classes becomes even more challenging \cite{liu2008exploratory}. The \textit{Streaming Ensemble Algorithm} (\textsc{sea}) \cite{street2001streaming} is used for classification drifting data streams using chunk-based learning for balanced data sets. The heuristic replacement technique is used to create the ensemble, so the model with the worst quality is removed. 

The following methods are suitable for imbalanced data with occurring concept drifts. The \textit{Learn++ for Concept Drift with \textsc{smote}} (Learn++.\textsc{cds}) \cite{ditzler2012incremental} is based on the Learn++.\textsc{nse} method \cite{elwell2011incremental}. Methods from Learn++ family use the weighted majority voting to make a final decision. \textsc{smote} is used to handle imbalanced data. For the \textit{Learn++ for Nonstationary and Imbalanced Environments} (Learn++.\textsc{nie}) \cite{ditzler2012incremental}, the ensemble is created with the penalty constraint in which the model is better when it performs both in the minority and majority class. Then bagging is used based on a subset of majority examples. The \textit{Over Under Sampling Ensemble} (\textsc{ouse}) \cite{gao2008classifying} employs oversampling to collect all previous minority class examples, as well as undersampling, which selects all majority class samples from the previous chunk. The \textit{Recursive Ensemble Approach} (\textsc{rea}) \cite{chen2011towards} uses selective accommodation of previous samples of the minority class for the current chunk based on the \textit{K-Nearest Neighbors}. The \textit{Kappa Updated Ensemble} (\textsc{kue}) \cite{cano2020kappa} uses the Kappa statistic rather than the \textit{Accuracy} for weighting and selection base models in the ensemble. The method ensures a high diversity and low complexity.
If the minority class is underrepresented, one-class classifiers can be used. One of the methods is \textit{One-Class Support Vector Machine Ensemble for Imbalanced data Stream} (\textsc{oceis}) \cite{klikowski2020employing}, which trains \textsc{ocsvm} for the majority and the minority classes based on clustered data.

\section{Proposition of the algorithm}
\label{sec:algorithm}

So far, not many methods dedicated to the classification of difficult data, such as non-stationary imbalanced data streams, have been developed. To overcome this challenging  data classification task, we propose the \textit{Hellinger Distance Weighted Ensemble} (\textsc{hdwe}). The method combines two approaches. Firstly, it employs the \textit{Accuracy Weighted Ensembles}, which is designed for drifting concepts. Secondly, the \textit{Hellinger Distance} is used to determine the weight in the ensemble based on the specific formula. This combination uses the advantage of both techniques. 

Usually, the classifier ensemble works better than a single classifier in the classification of \textit{concept drift} data streams. Therefore, \textsc{hdwe} bases on the \textsc{awe} method. Also, a similar mechanism for determining weights was used. The \textit{Hellinger Distance} presented in section \ref{sectionHD} is used as the classifier's weight in the ensemble. Thanks to this, \textsc{hdwe} can be resistant to the imbalance. Let us assume that the ensemble classifier with calculating weights for each model based on the \textit{Hellinger Distance} will be better for imbalanced data with the \textit{concept drift} than selected state-of-the-art methods.

The Algorithm \ref{algHDWE} outlines how the \textit{Hellinger Distance} (eq. \ref{eq:HD}) was used to the learning process of the ensemble. In the beginning, a pool of base classifiers $\Pi$ is empty. The data stream $DS$ is divided into equal chunks, except for the first initial one. The process of learning is conducted for each chunk $DS_i$. A new candidate classifier $m$ is trained based on the current chunk. Then the K-folds cross-validation is used to evaluate the model in the ensemble. The current chunk is split into K-parts (the default value of K is 5 for the \textsc{hdwe} method). The data is not shuffled. For each fold in cross-validation, the \textit{Hellinger Distance} function $HD$ calculates the score based on the candidate classifier and the fold from the current chunk using the eq. \ref{eq:HD}. The weight of the candidate $w_m$ is averaged through folds. The weights' normalization has no influence on results thus we do not implement it. In the next step, the \textit{Hellinger Distance} as a weight $w_k$ is calculated for each classifier $C_k$ in the ensemble. A new candidate is added to the ensemble. Afterward, if the ensemble's size is bigger than the given value $ ES $, the worst classifier is removed.

\begin{algorithm}
    \caption{Hellinger Distance Weighted Ensemble}
    \label{algHDWE}
    
    \textbf{Input:} \\
    \hspace*{\algorithmicindent} $HD$ -- the Hellinger Distance function\\
    \hspace*{\algorithmicindent} $DS$ -- the data stream \\
    \hspace*{\algorithmicindent} $DS_i$ -- $i$th chunk of the data stream $DS$ \\
    \hspace*{\algorithmicindent} $\Pi$ -- the pool of base classifiers \\
    \hspace*{\algorithmicindent} $C_k$ -- $k$th classifier of the pool of base classifiers $\Pi$ \\
    \hspace*{\algorithmicindent} $m$ -- the candidate classifier for the pool of base classifiers $\Pi$ \\
    \hspace*{\algorithmicindent} $w_m$ -- $m$th weight of the candidate classifier \\
    \hspace*{\algorithmicindent} $w_k$ -- $k$th weight of the $C_k$ classifier \\
    \hspace*{\algorithmicindent} $ES$ -- the given ensemble size \\
    
    \begin{algorithmic}[1]
        \STATE $\Pi \gets \varnothing $
        \FOR{each $DS_i \in DS$}
            \STATE Train classifier $m$ from $DS_i$
            \STATE $Scores \gets$ Compute $HD(m, DS_i)$ via cross validation using (\ref{eq:HD})
            \STATE $w_m \gets AverageScores$
            \FOR{each $C_k \in \Pi$}
                \STATE $w_k \gets$ Compute $HD(C_k, DS_i)$ using (\ref{eq:HD})
            \ENDFOR
            \STATE $\Pi \gets \Pi + m$
            \IF{$\abs{\Pi} > ES$}
                \STATE Remove the worst $C_k$ from $\Pi$
            \ENDIF
        \ENDFOR
    \end{algorithmic}
\end{algorithm}

The time complexity is based on the \textsc{awe} method \cite{wang2003mining}. Let $s$ be the size of the data set and $n$ -- the number of data chunks in the data stream. The complexity of building one classifier is $O(f(s))$. According to the Algorithm \ref{algHDWE}, the time complexity is equal $O(n \times f(\frac{s}{n}))$.

\section{Experimental evaluation}
\label{sec:experiments}

\subsection{Research goals}

The conducted experiments will try to answer the following research questions:

\begin{itemize}
    \item[RQ1:] How does the \textsc{hdwe} method work with different base classifiers?
    \item[RQ2:] Does the predictive performance of the \textsc{hdwe} outperform selected state-of-the-art methods?
    \item[RQ3:] How flexible is the \textsc{hdwe} in the non-stationary and dynamically imbalanced data sets?
\end{itemize}

\subsection{Setup}
\label{sec:setup}

The stream-learn module \cite{ksieniewicz2020stream} was used to conduct all experiments. It is a complete set of tools that helps to process data streams. First, the data streams were generated. Subsequently, \textit{Test-Then-Train} evaluator was used. The module allows the use of methods from the scikit-learn library \cite{scikitlearn} and has several classifiers and ensemble methods used as state-of-the-art methods. It also calculates selected metrics.

The project was implemented in the Python programming language. It uses also software such as SciPy \cite{2020SciPy_NMeth}, Pandas \cite{mckinney-proc-scipy-2010}, Numpy \cite{oliphant2006guide} for data processing and drawing charts. The project's implementation with this setup and results is available in the GitHub repository\footnote{https://github.com/joannagrzyb/HDWE}.

\subsubsection{Data sets}

Imbalanced data streams with the changing prior probabilities were used to conduct the experiments. The stream-learn module was employed to generate synthetic data. Table \ref{tab:strlearn} contains all attributes except the default ones. It was used to generate 84 data streams in total. 

The first five attributes are the same for each stream. As shown in the paper \cite{wang2003mining}, it is important to choose the right chunk size. When it is too big, the training time and error rate increase because the model cannot recognize the \textit{concept drift}. When the data chunk is small, the error also increases because it does not get enough training samples. After pre-experiments, the size of 500 objects in one data chunk was chosen for the research. 

The next attributes are variables, i.e., one value is selected from a row in the table. Streams generated in this way check many cases, especially different imbalanced levels from 1\% to 25\%. Two types of the imbalance were used: static and dynamic. The static imbalance does not change and the imbalance ratio is the same in the entire data stream . However, the dynamic imbalance changes over time with every new chunk. The initial minority class becomes the majority, and the majority becomes the minority. In two places among the number of chunks, the proportions of the classes are even. All streams contain 5 \textit{concept drifts} in two types: sudden and incremental. The sudden concept drift means that a state is promptly changed, and distribution is not adequate to the state. In the incremental concept drift, changes in the distribution are slower. According to \cite{krawczyk2017ensemble}, these are 2 of 3 main types of changes in the data stream. The random state ensures the replicability of generating the same streams.

\begin{table}[!ht]
\centering
\caption{Attributes of the generated data streams}
\label{tab:strlearn}
\scalebox{0.85}{
\begin{tabular}{|r|l|l|l|l|l|l|l|}
\hline
\multicolumn{1}{|c|}{\textbf{Attribute}}                         & \multicolumn{7}{c|}{\textbf{Value}}                                               \\ \hline
Number of samples                                                & \multicolumn{7}{l|}{100000}                                                       \\ \hline
Number of chunks                                                 & \multicolumn{7}{l|}{200}                                                          \\ \hline
Chunk size                                                       & \multicolumn{7}{l|}{500}                                                          \\ \hline
Number of classes                                                & \multicolumn{7}{l|}{2}                                                            \\ \hline
Number of features                                               & \multicolumn{7}{l|}{20 (15 informative + 5 redundant)}                            \\ \hline
Concept drifts                                                   & \multicolumn{3}{l|}{5 sudden}           & \multicolumn{4}{l|}{5 incremental}      \\ \hline
Random state                                                     & \multicolumn{2}{l|}{1111} & \multicolumn{2}{l|}{1234} & \multicolumn{3}{l|}{1567} \\ \hline
\begin{tabular}[c]{@{}r@{}}Stationary \\ imbalance\end{tabular}  & 1\%         & 3\%         & 5\%         & 10\%        & 15\%    & 20\%   & 25\%   \\ \hline
\begin{tabular}[c]{@{}r@{}}Dynamically \\ imbalance\end{tabular} & 1\%         & 3\%         & 5\%         & 10\%        & 15\%    & 20\%   & 25\%   \\ \hline
\end{tabular}}
\end{table}

Most of the experiments were run on computer-generated data streams, and the authors realize that better verification would be to use real data. However, we have to take into consideration the low availability of such data. According to analyzing possible real data streams, three benchmark streams were selected \cite{cano2020kappa}. Table \ref{tab:real} shows information about these data such as the imbalance ratio because each stream contains the concept drift, and after transformation they have the skew distribution. They were transformed into a 2-class by combining some classes. The data set \textit{covtype} contains information about the forest cover type and it has binary and normalized values. The original version has 7 classes but we merged first and the second class and compared them with the rest, so the classification problem became binary. Purpose of the second data set \textit{poker} is to predict poker hands composed of five playing cards. The data set has 10 classes, but we merged classes in the same way as in the previous data set. The aim of the \textit{insects} data set is classification of insects. To obtain the imbalanced data set and the binary problem, the second class has been selected and the rest has been merged. These streams' chunk size was 2000 to ensure that the model does not receive only one class during classification.

\begin{table}[!ht]
\centering
\caption{Properties of real data sets}
\scalebox{0.85}{
\begin{tabular}{|l|l|l|l|}
\hline
\textbf{Data set}    & \textbf{Instances} & \textbf{Attributes} & \textbf{Imbalance Ratio} \\ \hline
covtypeNorm-1-2vsAll & 267 000                      & 54                            & 25\%                        \\ \hline
poker-lsn-1-2vsAll   & 360 000                      & 10                            & 10\%                       \\ \hline
2vsA\_INSECTS   & 355 274                      & 33                            & 23\%                       \\ \hline
\end{tabular}}
\label{tab:real}
\end{table}

\subsubsection{Used classifiers}
\label{sec:methods}

In the first experiment, the following base classifiers were used. \textsc{hddt} was described in details in section \ref{sec:relatedworks}. The other methods were used from the scikit-learn library.

\begin{itemize}
    \item \textit{Gaussian Naive Bayes} \cite{zhang2004ithe} -- \textsc{gnb} 
    \item \textit{Multi-Layer Perceptron classifier} \cite{hampshire1991equivalence} -- \textsc{mlp}
    \item \textit{Decision Tree classifier} \cite{steinberg2009cart} -- \textsc{cart}
    \item \textit{Hellinger Distance Decision Tree} \cite{cieslak2012hellinger} -- \textsc{hddt}
    \item \textit{K-Nearest Neighbors classifier} \cite{goldberger2005neighbourhood} -- \textsc{knn}
    \item \textit{C-Support Vector Classification} \cite{chang2011libsvm} -- \textsc{svc}
\end{itemize}

The second experiment contains a comparison between \textsc{hdwe} and undermentioned ensembles. \textsc{sea} and \textsc{awe} was used from the stream-learn module. 

\begin{itemize}
    \item \textit{Streaming Ensemble Algorithm} \cite{street2001streaming} -- \textsc{sea}
    \item \textit{Accuracy Weighted Ensemble} \cite{wang2003mining} -- \textsc{awe}
    \item \textit{Learn++ for Concept Drift with \textsc{smote}} \cite{ditzler2012incremental} -- Learnpp\textsc{cds}
    \item \textit{Learn++ for Nonstationary and Imbalanced Environments} \cite{ditzler2012incremental} -- Learnpp\textsc{nie}
    \item \textit{Over and Under Sampling Ensemble} \cite{gao2008classifying} -- \textsc{ouse}
    \item \textit{Recursive Ensemble Approach} \cite{chen2011towards} -- \textsc{rea}
\end{itemize}

During experiments, the number of estimators used in each ensemble was 10, because according to \cite{zenobi2001using} the number of models in the ensemble should be between 10 and 50 to strike a balance between \textit{accuracy} and \textit{diversity}. Table \ref{tab:attrens} shows values of other parameters.

\begin{table}[!ht]
\centering
\caption{Attributes of the ensemble classifiers}
\label{tab:attrens}
\scalebox{0.85}{
\begin{tabular}{|l|r|l|}
\hline
\multicolumn{1}{|c|}{\textbf{Method}} & \multicolumn{1}{c|}{\textbf{Attribute}} & \multicolumn{1}{c|}{\textbf{Value}} \\ \hline
\textsc{sea}                                   & Metric                                  & Accuracy                            \\ \hline
\textsc{awe}                                   & Number of folds in cross validation     & 5                                   \\ \hline
\multirow{2}{*}{\begin{tabular}[c]{@{}l@{}}Learnpp\textsc{cds}\\ Learnpp\textsc{nie}\end{tabular}}          & Parameter a                             & 2                                   \\ \cline{2-3} 
                                      & Parameter b                             & 2                                   \\ \hline
\textsc{ouse}                                  & Number of chunks                        & 10                                  \\ \hline
\textsc{rea}                                   & Balance ratio                           & 0,5                                 \\ \hline
\end{tabular}}
\end{table}

\subsubsection{Metrics}
\label{sec:metrics}

In this work, we will focus on the binary classification task. Before presenting the metrics, let us define a confusion matrix (Table \ref{tab:confmatrix}) \cite{wozniak2013hybrid}. \textsc{tp} (\emph{true positive}) stands for the number of correctly classified positive instances (minority class).  \textsc{tn} (\emph{true negative}) – correctly classified negative examples (majority class). \textsc{fp} (\emph{false positive}) and \textsc{fn} (\emph{false negative}) are the numbers of incorrectly classified objects from positive and negative classes, respectively.

\begin{table}[!ht]
\caption{Confusion matrix for binary problem}
\label{tab:confmatrix}
\centering
\scalebox{0.85}{
\begin{tabular}{llcc}
                                                                                                     &                                   & \multicolumn{2}{c}{Actual values}                                                                                                                                                     \\ \cline{3-4} 
                                                                                                     & \multicolumn{1}{l|}{}             & \multicolumn{1}{l|}{Positive (1)}                                                         & \multicolumn{1}{l|}{Negative (0)}                                                         \\ \cline{2-4} 
\multicolumn{1}{c|}{\multirow{2}{*}{\begin{tabular}[c]{@{}c@{}}Predicted \\ values\end{tabular}}} & 
\multicolumn{1}{l|}{Positive (1)} &
\multicolumn{1}{c|}{\begin{tabular}[c]{@{}c@{}}\textsc{tp} \\ True Positive\end{tabular}} &
\multicolumn{1}{c|}{\begin{tabular}[c]{@{}c@{}}\textsc{fp}\\ False Positive\end{tabular}} \\ \cline{2-4} 
\multicolumn{1}{c|}{}                                                                                & \multicolumn{1}{l|}{Negative (0)} & \multicolumn{1}{c|}{\begin{tabular}[c]{@{}c@{}}\textsc{fn}\\ False Negative\end{tabular}} & \multicolumn{1}{c|}{\begin{tabular}[c]{@{}c@{}}\textsc{tn}\\ True Negative\end{tabular}}  \\ \cline{2-4} 
\end{tabular}}
\end{table}

On the basis of the confusion matrix, the following metrics can be calculated that show the quality of the classification \cite{alpaydin2009introduction} \cite{wozniak2013hybrid}:

\begin{itemize}
    \item \textit{Accuracy} is one of the most popular threshold metric. This is a measurement of the correct predictions between all predicted and actual values. 
    \begin{equation}
        Accuracy = \frac{TP+TN}{TP+TN+FP+FN}
    \end{equation}
    \item \textit{Recall} (other name \textit{Sensitivity} or \textit{True Positive Rate} -- \textsc{tpr}) informs with what quality the model recognizes objects as the minority class, which actually belong to this class.
    \begin{equation}
    Recall = \frac{TP}{TP+FN}
    \end{equation}
    \item \textit{Specificity} (also known as \textit{True Negative Rate} -- \textsc{tnr}) returns the quality of the model as it correctly assigned the objects to the majority class in relation to how many objects in that class are.
    \begin{equation}
    Specificity = \frac{TN}{TN+FP}
    \end{equation}
    \item \textit{Precision}
    \begin{equation}
    Precision = \frac{TP}{TP+FP}
    \end{equation}
    \item \textit{$F_1$ score}
    \begin{equation}
    F_1score = 2 \times \frac{Precision \times Recall}{Precision + Recall}
    \end{equation}
    \item \textit{Balanced Accuracy} (\textsc{bac})
    \begin{equation}
    \textsc{bac} = \frac{Recall + Specificity}{2}
    \end{equation}
    \item \textit{Geometric mean score} (G--mean)
    \begin{equation}
    G-mean = \sqrt{Recall \times Specificity}
    \end{equation}
    \item \textit{False Positive Rate} (\textsc{fpr})
    \begin{equation}
    FPR = 1 - Specificity = \frac{FP}{TN+FP}
    \end{equation}
    \item \textit{False Negative Rate} (\textsc{fnr})
    \begin{equation}
    FNR = 1 - Recall = \frac{FN}{TP+FN}
    \end{equation}
\end{itemize}

For imbalanced data sets, the \textit{Accuracy} is inappropriate because it may favor the majority class. Thus, more appropriate metrics for this type of data sets are those that measure only one class, such as \textit{Recall} or \textit{Specificity}. A combination of these metrics can also be useful, e.g. \textit{$F_1$ score}, \textit{G--mean} and \textit{Balanced Accuracy}.

\subsubsection{Statistical tests}





The statistical analysis may improve the readability of results and strengthen conclusions from experiments. One of the acclaimed work on analysis methods is \cite{demvsar2006statistical}, where Demsar showed different statistical tests that may be performed depending on the number of classifiers and the number of data sets. In our research the nonparametric \textit{Friedman test} with the \textit{Nemenyi post-hoc test} are used.

\subsection{Results}

This section presents an analysis of the results of the conducted experiments. The first of them check the base classifier type's impact on the \textsc{hdwe} method. Then two base classifiers \textsc{svc} and \textsc{hddt} were selected. For each of them, the \textsc{hdwe} method was compared with selected state-of-the-art methods. The last part shows the dependence of the imbalance.

\subsubsection{Experiment 1 -- base classifiers for \textsc{hdwe}}

The first experiment compares base classifiers in the \textsc{hdwe} method with each other. As described precisely in section \ref{sec:setup}, 6 base classifiers were compared in 84 generated data streams. 

\begin{figure}[!ht]
\centering
\begin{subfigure}{0.8\textwidth}
\includegraphics[width=\linewidth]{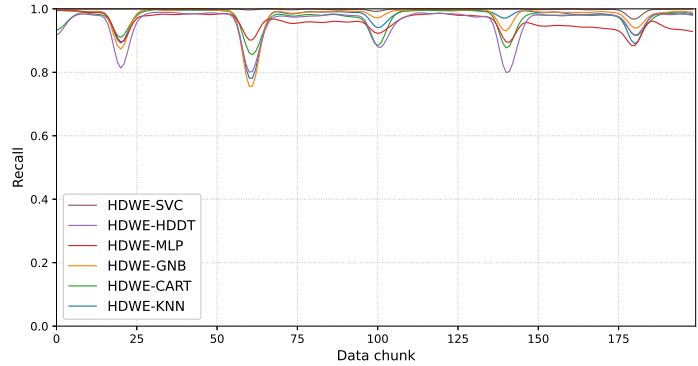}
\caption{Recall} \label{fig:experiment1sudA}
\end{subfigure}
\begin{subfigure}{0.8\textwidth}
\includegraphics[width=\linewidth]{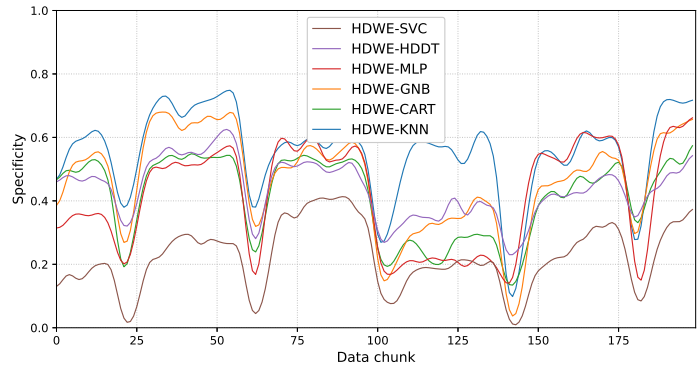}
\caption{Specificity} \label{fig:experiment1sudB}
\end{subfigure}
\caption{Performance scores for base classifiers -- the generated data stream with the sudden concept drift and the stationary imbalance} \label{fig:experiment1sud}
\end{figure}

\begin{figure}[!ht]
\centering
\begin{subfigure}{0.8\textwidth}
\includegraphics[width=\linewidth]{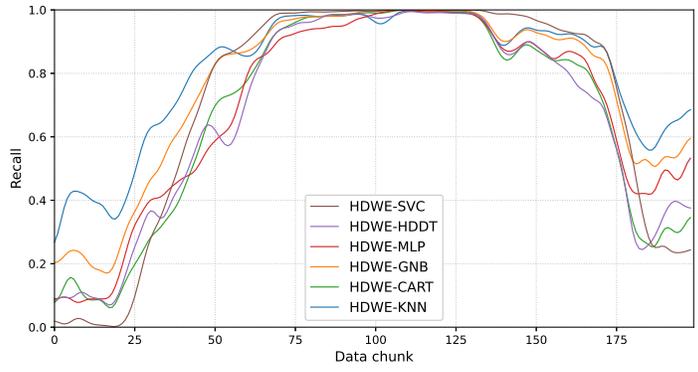}
\caption{Recall} \label{fig:experiment1incA}
\end{subfigure}
\begin{subfigure}{0.8\textwidth}
\includegraphics[width=\linewidth]{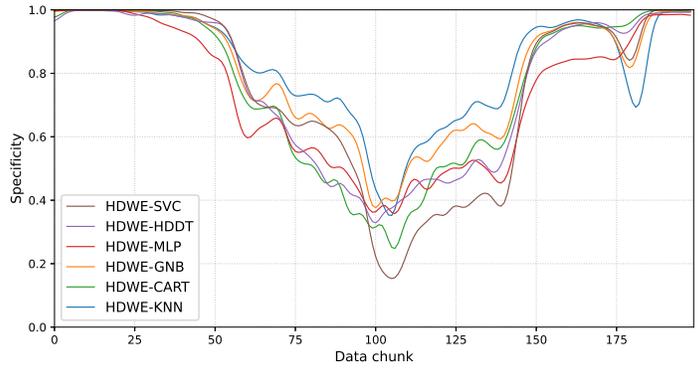}
\caption{Specificity} \label{fig:experiment1incB}
\end{subfigure}
\caption{Performance scores for base classifiers -- the generated data stream with the incremental concept drift and the dynamically imbalance} \label{fig:experiment1inc}
\end{figure}

Four example graphs of measured quality \textit{Recall} and \textit{Specificity} are shown in Figures \ref{fig:experiment1sud} and \ref{fig:experiment1inc}. Both data streams have 10\% of the imbalance. Figures \ref{fig:experiment1sudA} and \ref{fig:experiment1sudB} show the sudden \textit{concept drift} with the stationary imbalance. Figure \ref{fig:experiment1sudA} clearly shows five \textit{concept drifts}. In addition, all methods achieve the highest \textit{Recall} value, which means in this case that the minority class has been well recognized. Figure \ref{fig:experiment1sudB} shows the significant difference in the level of \textit{Specificity} for different methods, so the \textsc{hdwe} method is worse at recognizing the majority class.

In Figures \ref{fig:experiment1incA} and \ref{fig:experiment1incB} it is the incremental \textit{concept drift} with the dynamic imbalance. It looks a little different for the dynamic \textit{concept drift} and the imbalance in Figures \ref{fig:experiment1incA} and \ref{fig:experiment1incB}. Both figures show that the methods reach maximum values, but after some time the quality decreases.

\begin{figure}[!ht]
\centering
\begin{subfigure}{0.49\textwidth}
\includegraphics[width=\linewidth]{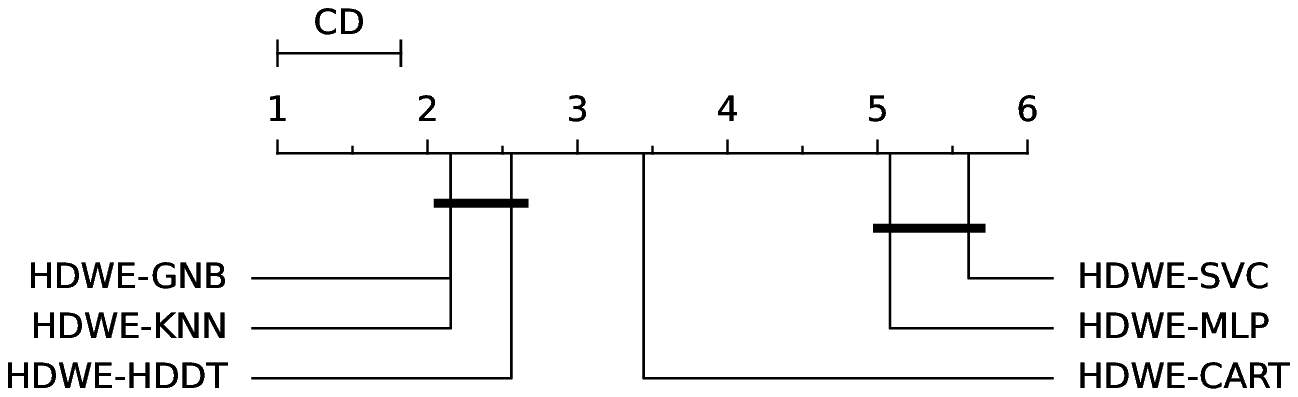}
\caption{Balanced Accuracy} \label{fig:ex1_NemenyiA}
\end{subfigure}
\begin{subfigure}{0.49\textwidth}
\includegraphics[width=\linewidth]{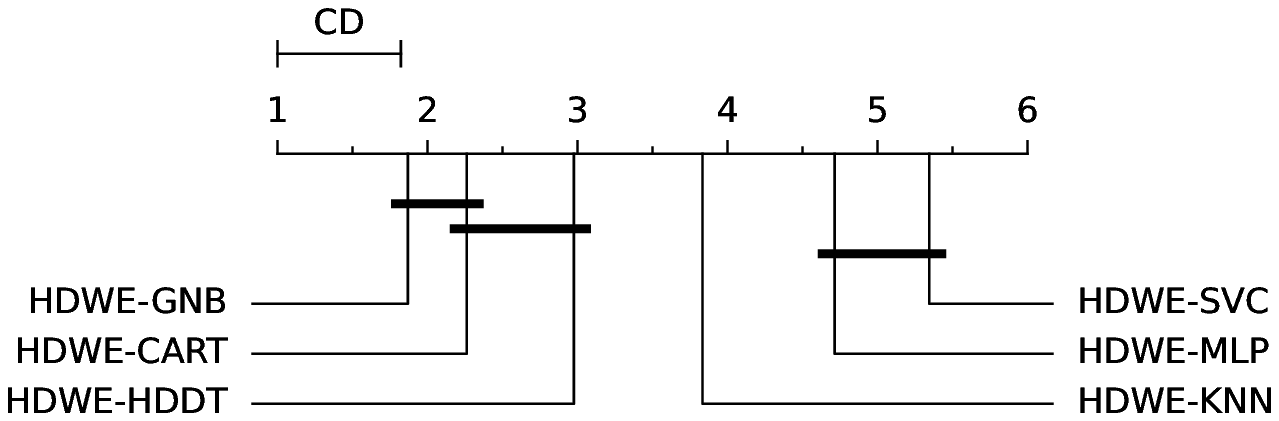}
\caption{$F_1$ score} \label{fig:ex1_NemenyiB}
\end{subfigure}
\begin{subfigure}{0.49\textwidth}
\includegraphics[width=\linewidth]{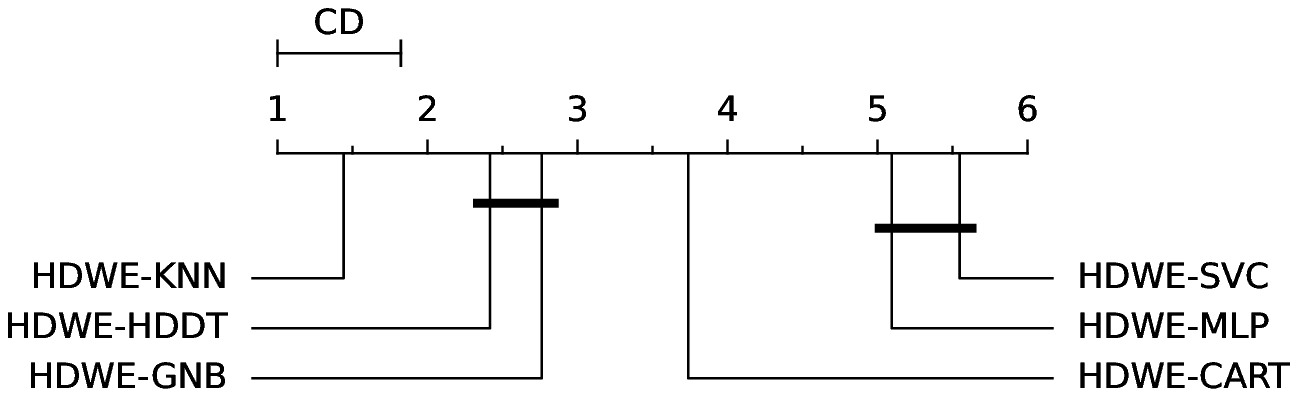}
\caption{G--mean} \label{fig:ex1_NemenyiC}
\end{subfigure}
\begin{subfigure}{0.49\textwidth}
\includegraphics[width=\linewidth]{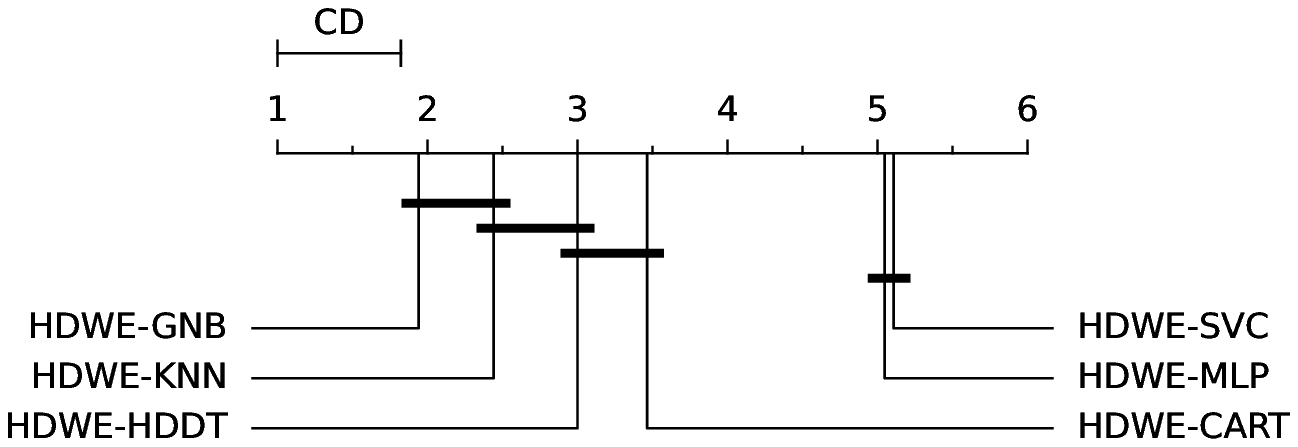}
\caption{Precision} \label{fig:ex1_NemenyiD}
\end{subfigure}
\begin{subfigure}{0.49\textwidth}
\includegraphics[width=\linewidth]{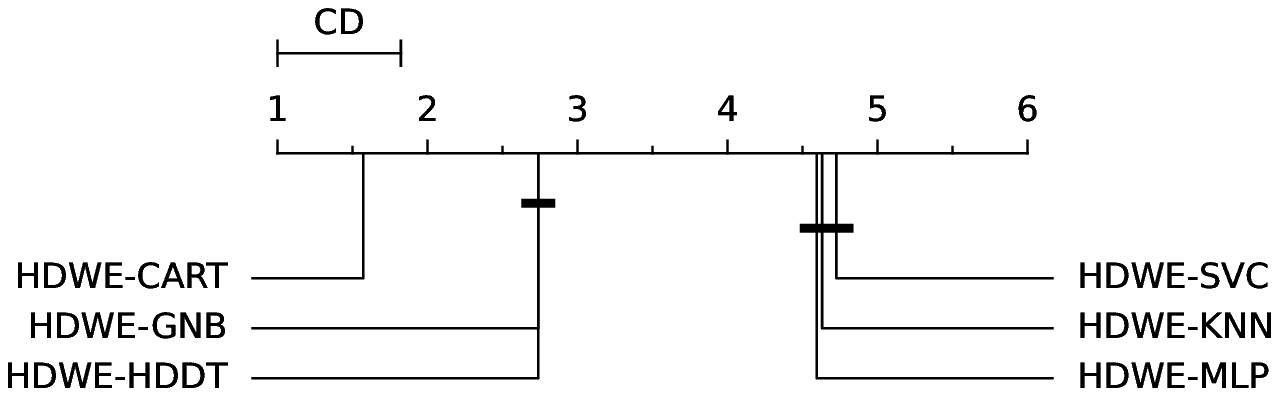}
\caption{Recall} \label{fig:ex1_NemenyiE}
\end{subfigure}
\begin{subfigure}{0.49\textwidth}
\includegraphics[width=\linewidth]{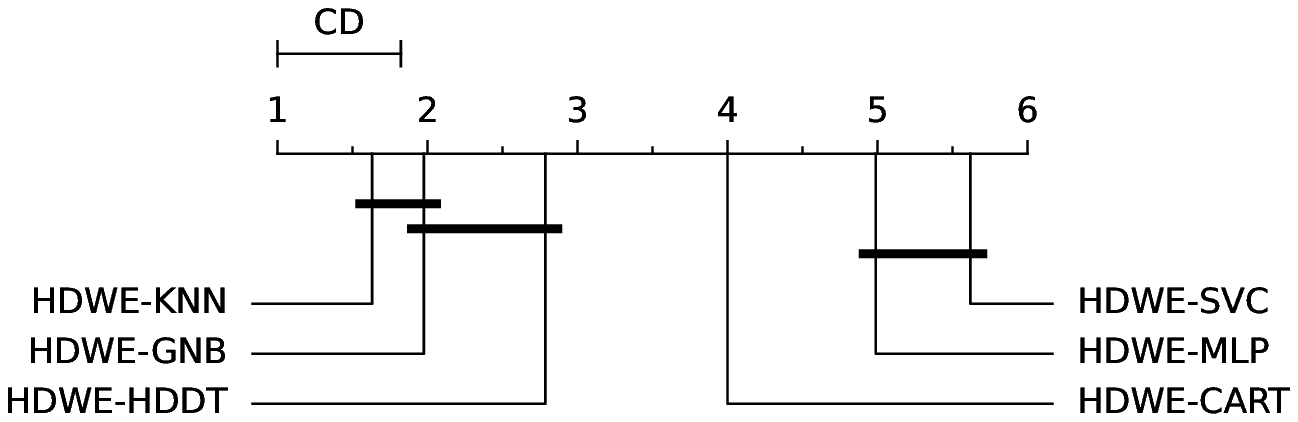}
\caption{Specificity} \label{fig:ex1_NemenyiF}
\end{subfigure}
\caption{Diagrams of critical difference for Nemenyi test for base classifiers} \label{fig:ex1-1_Nemenyi}
\end{figure}

Based on the selected figures from two streams, it is impossible to determine which method is statistically the best. Thus, each stream's results were averaged, and then the average rankings for each method and six metrics were calculated separately. To better compare all methods and find these statistically different, the \textit{Friedman test} is performed \cite{demvsar2006statistical}. For the p-value equal 0.05, the hypothesis H0 is rejected. Once the H0 hypothesis is rejected, it can proceed to the \textit{Nemenyi test} \cite{demvsar2006statistical}. Figure \ref{fig:ex1-1_Nemenyi} shows diagrams for every metric calculated based on the \textit{Nemenyi test}. It is a post-hoc test. An axis represents the ranks of the method, where the higher number means the better method. The best methods are on the right side of every diagram. The critical difference is calculated for the confidence level $\alpha=0.05$, average ranks of scores in all generated data streams. Methods that are not significantly different are connected with a thick horizontal line. The other ones are significantly different, which means that its average ranks differ at least a critical difference.

Based on the above analyzes, it could be concluded that the base classifier \textsc{svc} in the \textsc{hdwe} method is statistically significantly better than the other. Except for \textsc{mlp}, which is not significantly different from \textsc{svc}.

\subsubsection{Experiment 2 -- comparison with state-of-the-art methods with the base classifier: \textsc{svc}}

This section presents the second experiment. It contains a comparison between the \textsc{hdwe} method with state-of-the-art methods: \textsc{sea}, \textsc{awe}, Learn++.\textsc{cds}, Learn++.\textsc{nie}, \textsc{ouse}, \textsc{rea}. All of these methods are ensembles, which means they must use the base classifier. \textsc{svc} was chosen for the study first because it proved to be statistically the best in the first experiment. 

\begin{figure}[!ht]
\centering
\begin{subfigure}{0.8\textwidth}
\includegraphics[width=\linewidth]{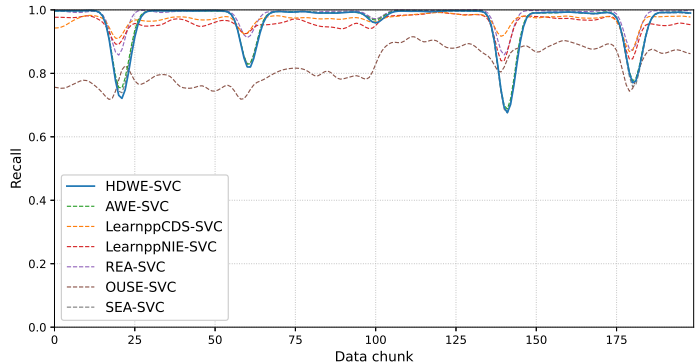}
\caption{Recall} \label{fig:experiment2asudA}
\end{subfigure}
\begin{subfigure}{0.8\textwidth}
\includegraphics[width=\linewidth]{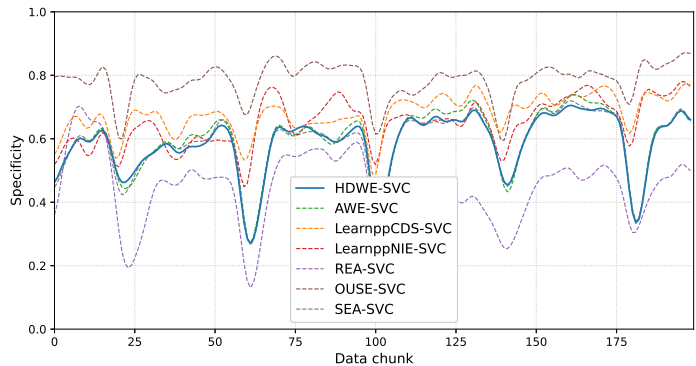}
\caption{Specificity} \label{fig:experiment2asudB}
\end{subfigure}
\caption{Performance scores for state-of-the-art methods with base classifier \textsc{svc} -- the generated data stream with the sudden concept drift and the stationary imbalance} \label{fig:experiment2asud}
\end{figure}

\begin{figure}[!ht]
\centering
\begin{subfigure}{0.8\textwidth}
\includegraphics[width=\linewidth]{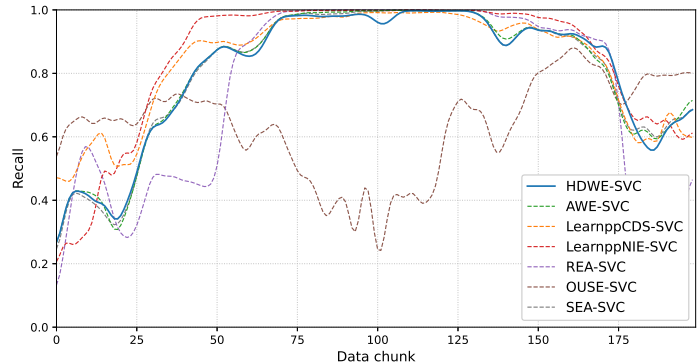}
\caption{Recall} \label{fig:experiment2aincA}
\end{subfigure}
\begin{subfigure}{0.8\textwidth}
\includegraphics[width=\linewidth]{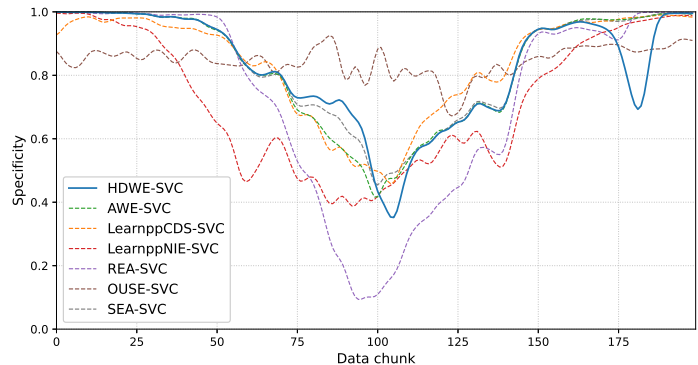}
\caption{Specificity} \label{fig:experiment2aincB}
\end{subfigure}
\caption{Performance scores for state-of-the-art methods with base classifier \textsc{svc} -- the generated data stream with the incremental concept drift and the dynamically imbalance} \label{fig:experiment2ainc}
\end{figure}

From all generated data streams, two examples were selected with the imbalance ratio 10\%, and two metrics: \textit{Recall} -- indicating the minority class and \textit{Specificity} -- indicating the majority class. Figures \ref{fig:experiment2asudA} and \ref{fig:experiment2asudB} show five sudden \textit{concept drifts} and the stationary imbalance. The \textsc{hdwe} method is working just as well as the other. It is worth noting that the \textsc{ouse} method is better when recognizing the majority class. Figures \ref{fig:experiment2aincA} and \ref{fig:experiment2aincB} show five incremental \textit{concept drifts} and the dynamical imbalance.

\begin{figure}[!ht]
\centering
\begin{subfigure}{0.49\textwidth}
\includegraphics[width=\linewidth]{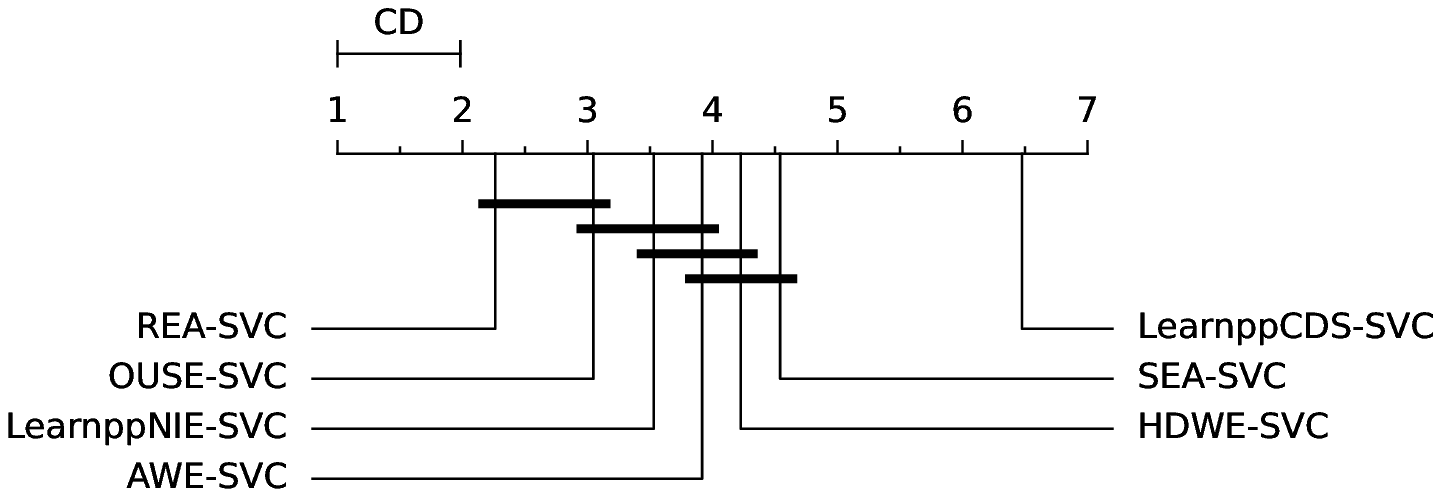}
\caption{Balanced Accuracy} \label{fig:ex2a_NemenyiA}
\end{subfigure}
\begin{subfigure}{0.49\textwidth}
\includegraphics[width=\linewidth]{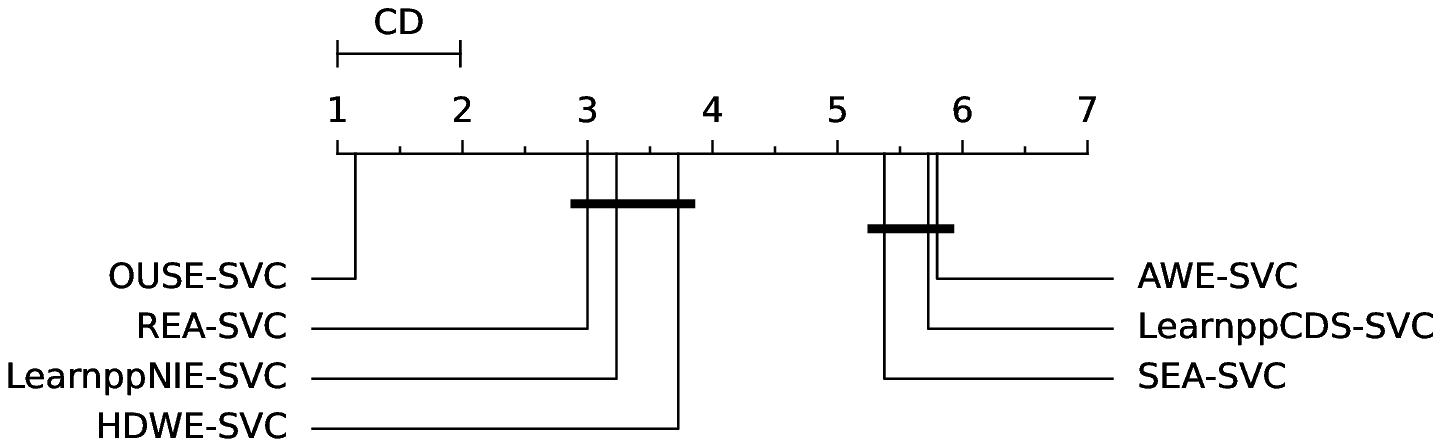}
\caption{$F_1$ score} \label{fig:ex2a_NemenyiB}
\end{subfigure}
\begin{subfigure}{0.49\textwidth}
\includegraphics[width=\linewidth]{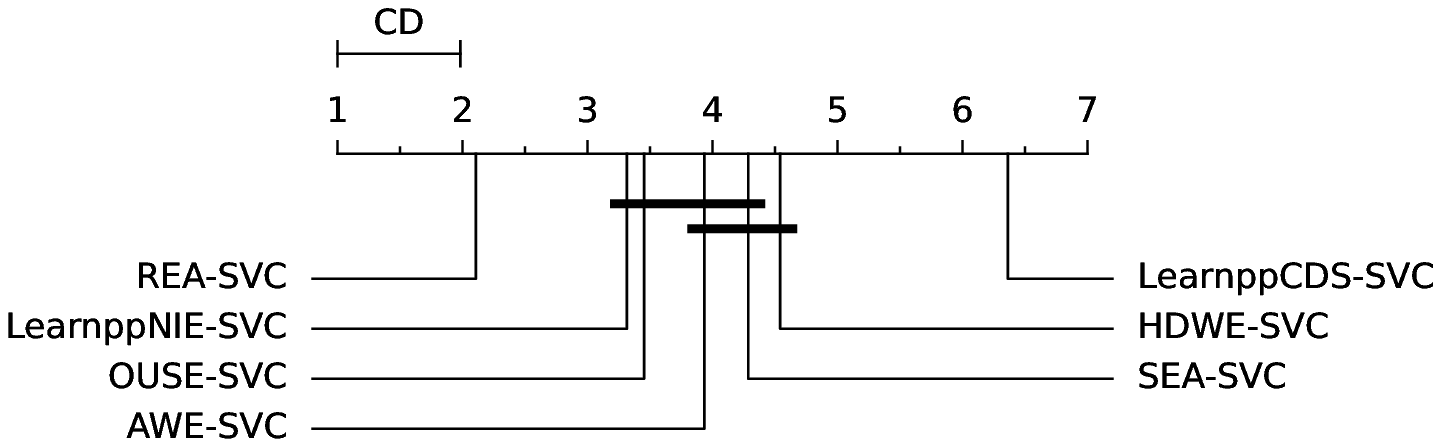}
\caption{G--mean} \label{fig:ex2a_NemenyiC}
\end{subfigure} 
\begin{subfigure}{0.49\textwidth}
\includegraphics[width=\linewidth]{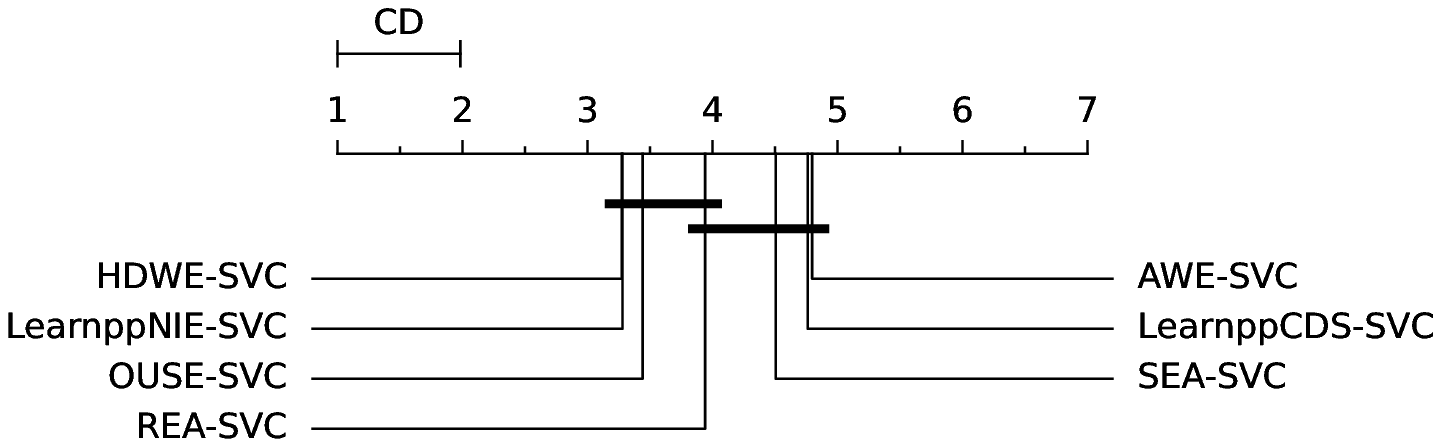}
\caption{Precision} \label{fig:ex2a_NemenyiD}
\end{subfigure}
\begin{subfigure}{0.49\textwidth}
\includegraphics[width=\linewidth]{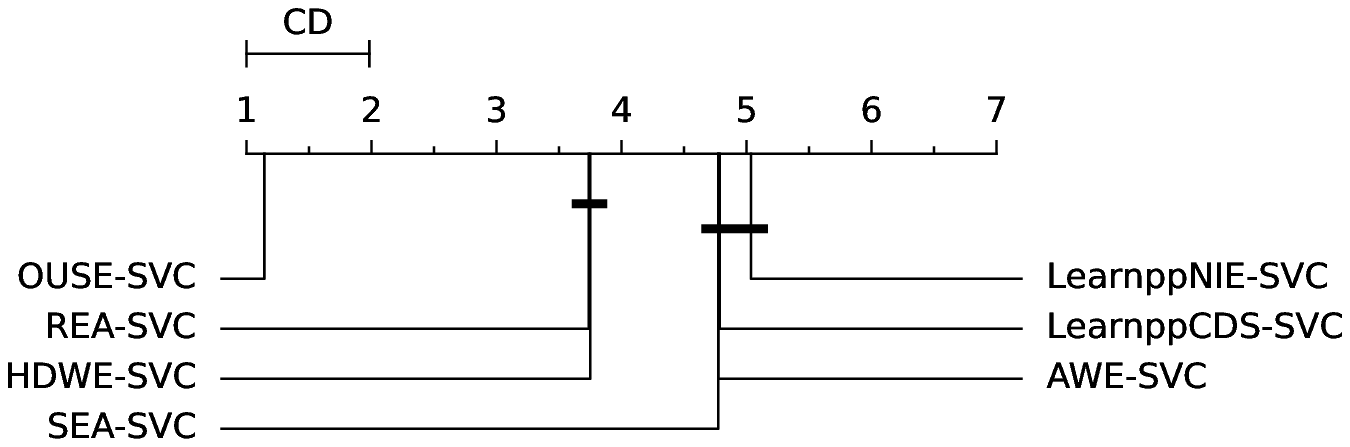}
\caption{Recall} \label{fig:ex2a_NemenyiE}
\end{subfigure}
\begin{subfigure}{0.49\textwidth}
\includegraphics[width=\linewidth]{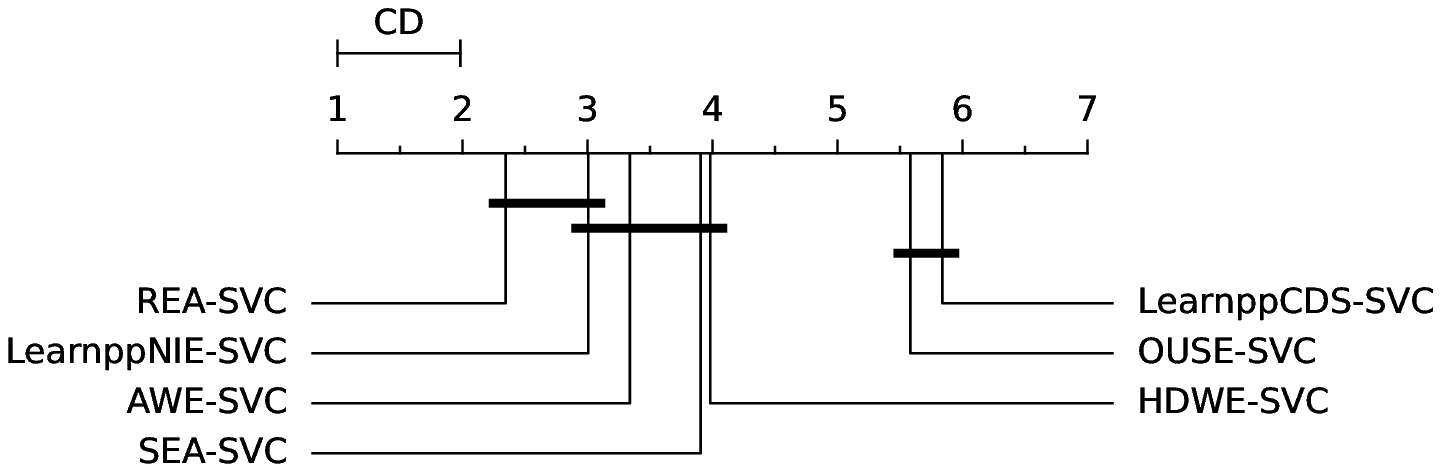}
\caption{Specificity} \label{fig:ex2a_NemenyiF}
\end{subfigure}
\caption{Diagrams of critical difference for Nemenyi test for state-of-the-art methods with the base classifier \textsc{svc}}
\label{fig:ex2a-1_Nemenyi}
\end{figure}

As in the first experiment, the average ranks of all methods were calculated. The \textit{Friedman test} is carried out. For p-value equal 0.05, the hypothesis H0 is rejected, so the \textit{Nemenyi test} can be conducted. Figure \ref{fig:ex2a-1_Nemenyi} shows the \textit{Nemenyi test}. The axis represents the ranks of the methods. The better methods are on the right. The confidence level is $\alpha=0.05$ for calculating \textsc{cd}. The thick, horizontal line connects methods which are not significantly different.

Depending on the metrics, the best results were obtained by Learn++.\textsc{cds}, \textsc{awe}, and Learn++.\textsc{nie}. For half of the metrics (\textit{Balanced Accuracy}, \textit{G--mean}, \textit{Specificity}), the \textsc{hdwe} method is on the right, which means it is quite good compared to others. For example, for \textit{Recall}, it is significantly better than the \textsc{ouse} method.

\begin{figure}[!ht]
\centering
\includegraphics[width=0.9\linewidth]{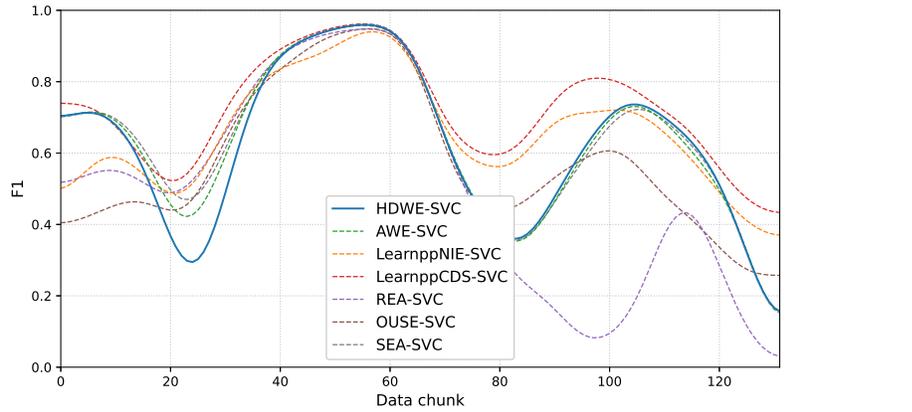}
\caption{$F_1$ score for state-of-the-art methods with base classifier \textsc{svc} -- real data stream (\textit{covtype})} \label{fig:ex4_SVC_covtype}
\end{figure}
\begin{figure}[!ht]
\centering
\includegraphics[width=0.9\linewidth]{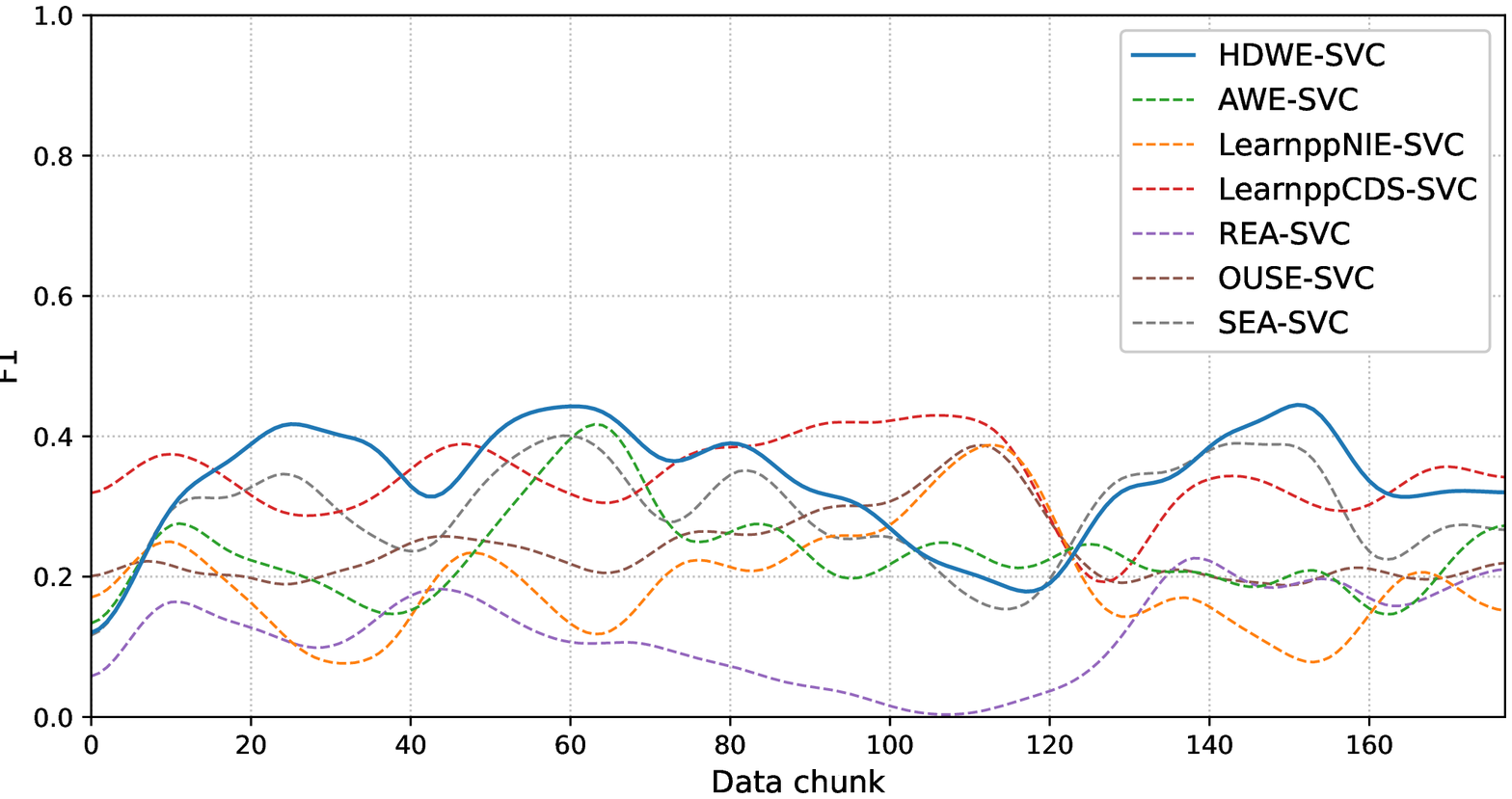}
\caption{$F_1$ score for state-of-the-art methods with base classifier \textsc{svc} -- real data stream (\textit{poker})} \label{fig:ex4_SVC_poker}
\end{figure}
\begin{figure}[!ht]
\centering
\includegraphics[width=0.9\linewidth]{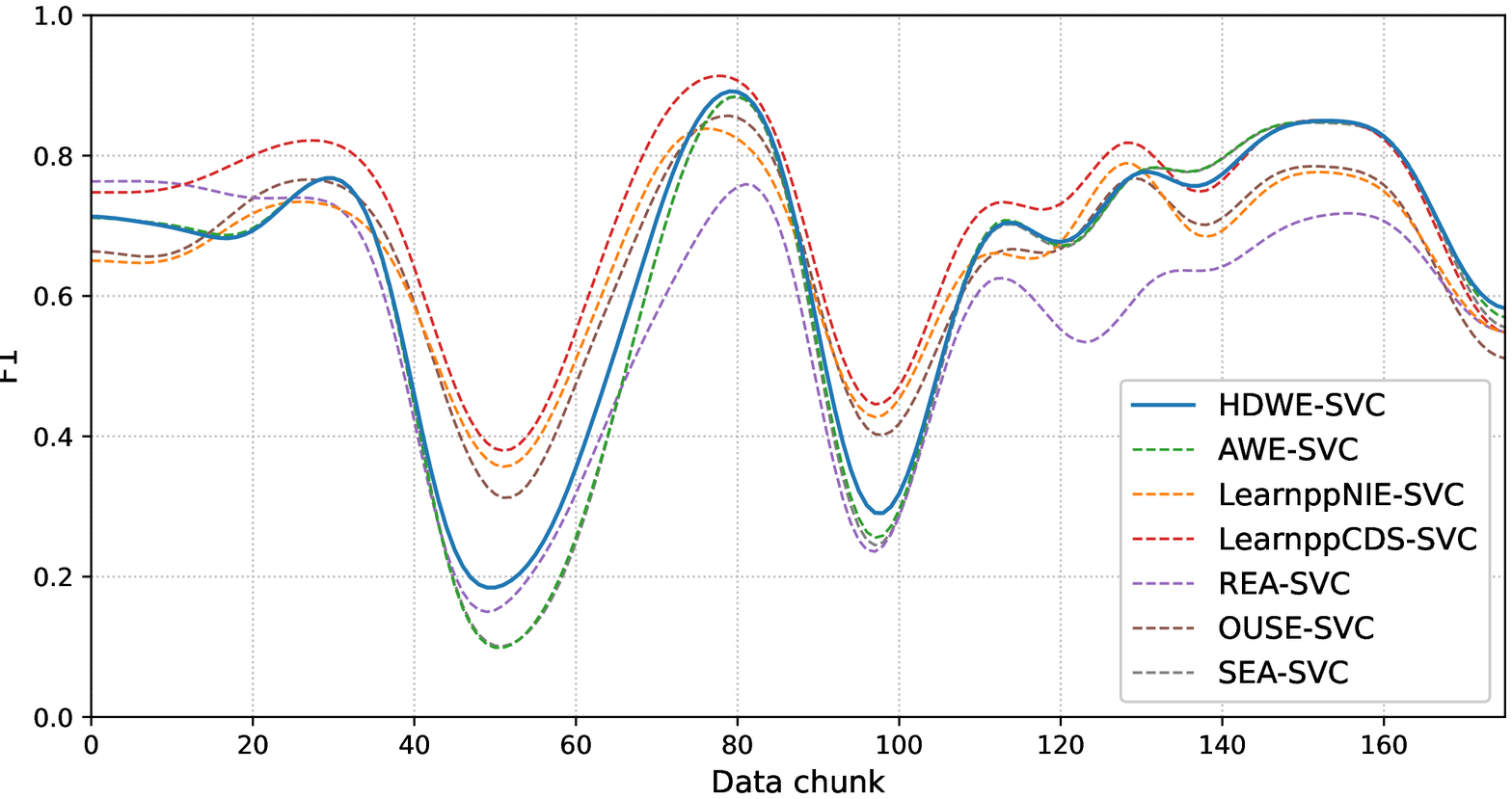}
\caption{$F_1$ score for state-of-the-art methods with base classifier \textsc{svc} -- real data stream (\textit{insects})} \label{fig:ex4_SVC_insects}
\end{figure}
\begin{figure}[!ht]
\centering
\begin{subfigure}{0.48\textwidth}
\includegraphics[width=\linewidth]{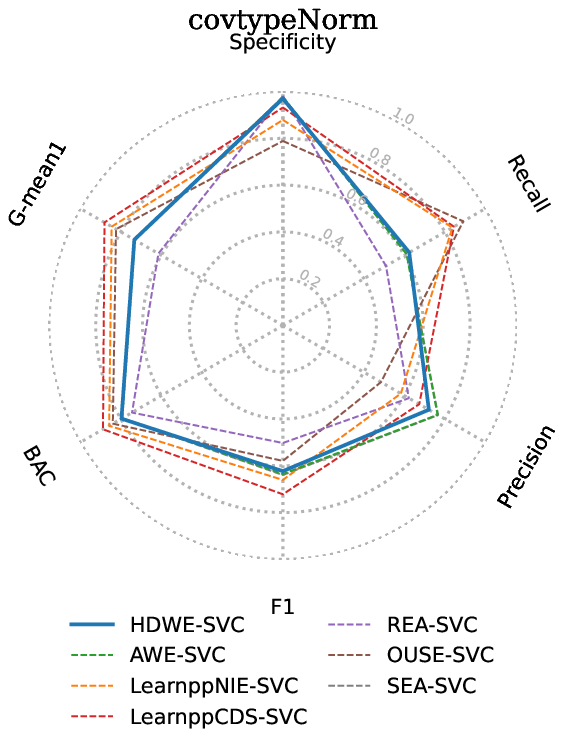}
\caption{\textit{Covtype} data set} \label{fig:ex4_radar_SVC_A}
\end{subfigure} \hfill
\begin{subfigure}{0.48\textwidth}
\includegraphics[width=\linewidth]{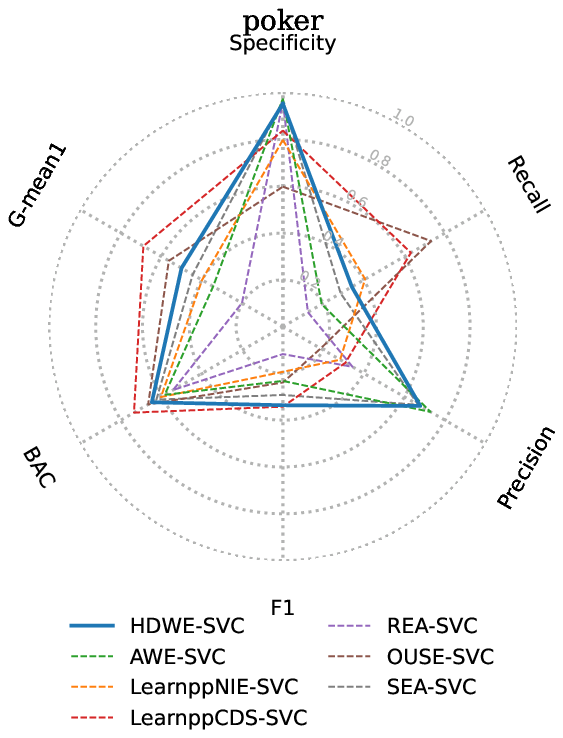}
\caption{\textit{Poker} data set} \label{fig:ex4_radar_SVC_B}
\end{subfigure} \hfill
\begin{subfigure}{0.48\textwidth}
\includegraphics[width=\linewidth]{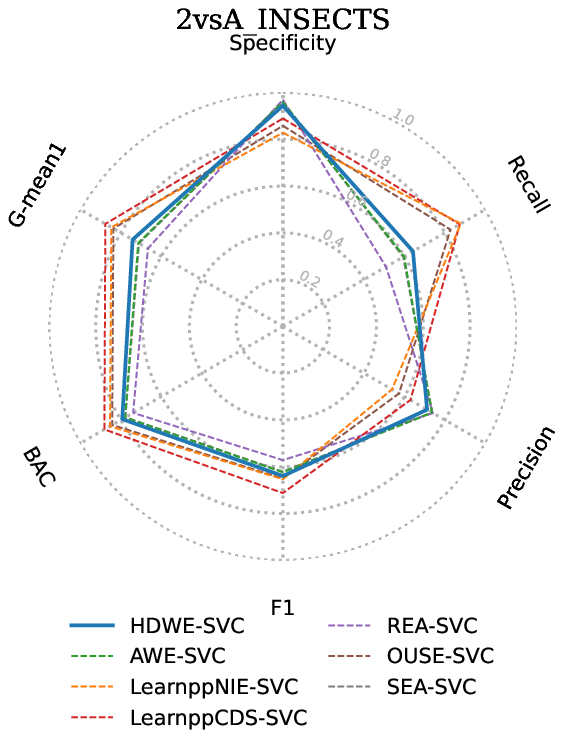}
\caption{\textit{Insects} data set} \label{fig:ex4_radar_SVC_ii}
\end{subfigure}
\caption{Performance scores for state-of-the-art methods with base classifier \textsc{svc} -- real data sets} \label{fig:ex4_radar_SVC}
\end{figure}

Figures \ref{fig:ex4_SVC_covtype}, \ref{fig:ex4_SVC_poker} and \ref{fig:ex4_SVC_insects} show $F_1$ score for all methods tested on the real data sets \textit{covtype}, \textit{poker} and \textit{insects} respectively. Figure \ref{fig:ex4_radar_SVC} includes radar representation of all metrics. For each case, \textsc{hdwe} has reasonable values of \textit{G--mean}, $F_1$ score and \textit{Balanced Accuracy} compared to other methods. It achieves the highest \textit{Specificity} and \textit{Precision} and lower \textit{Recall} than others. For all methods a high value of \textit{Specificity} may lead to overfitting toward the majority class.

\subsubsection{Experiment 2 -- comparison with state-of-the-art methods with the base classifier: \textsc{hddt}}

The second base classifier chosen to conduct a similar analysis of the result is \textsc{hddt}. It is a decision tree that returns interesting results because, like the \textsc{hdwe} method, it bases on the \textit{Hellinger Distance}. 

\begin{figure}[!ht]
\centering
\begin{subfigure}{0.8\textwidth}
\includegraphics[width=\linewidth]{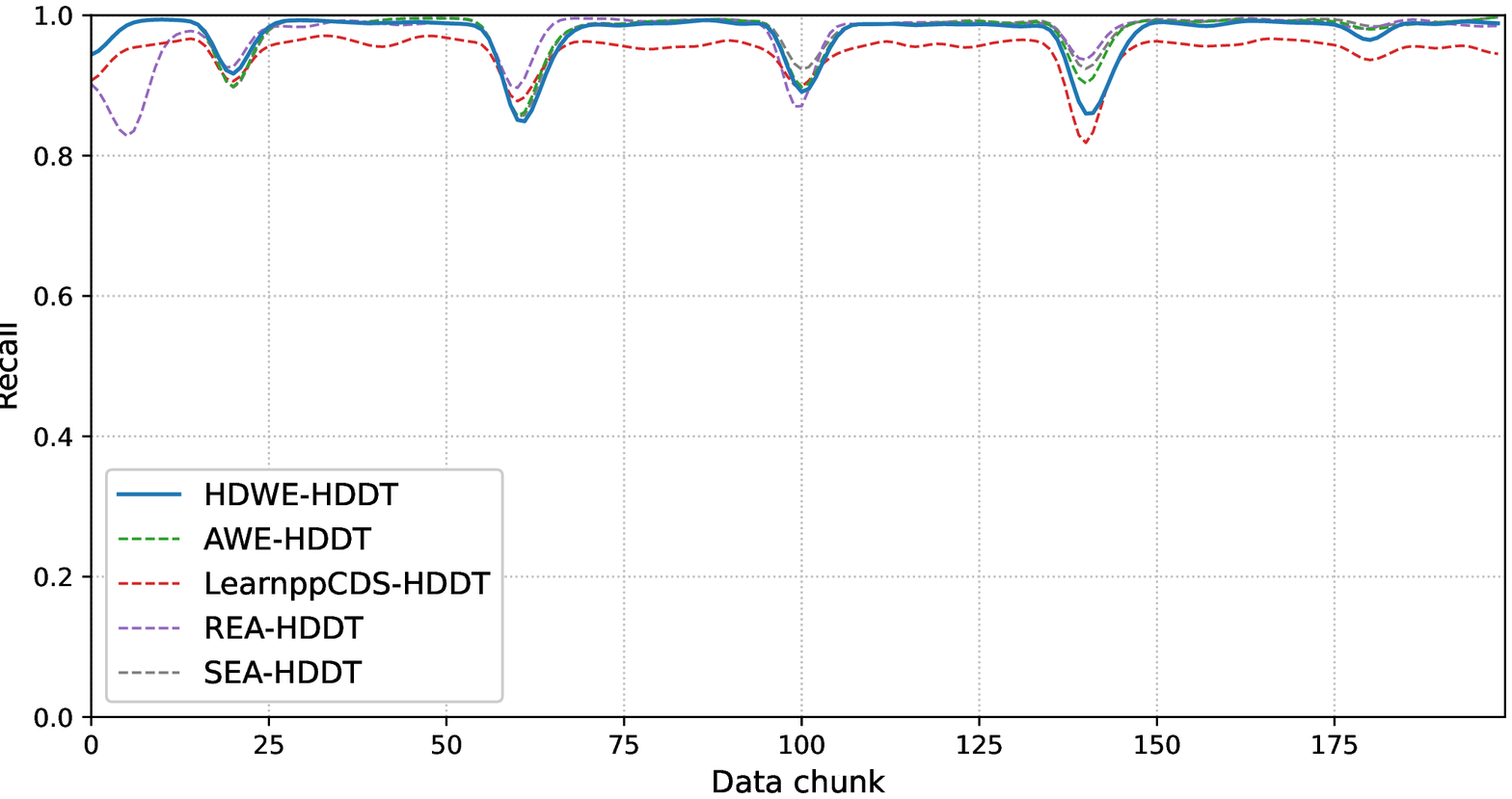}
\caption{Recall} \label{fig:experiment2bsudA}
\end{subfigure}
\begin{subfigure}{0.8\textwidth}
\includegraphics[width=\linewidth]{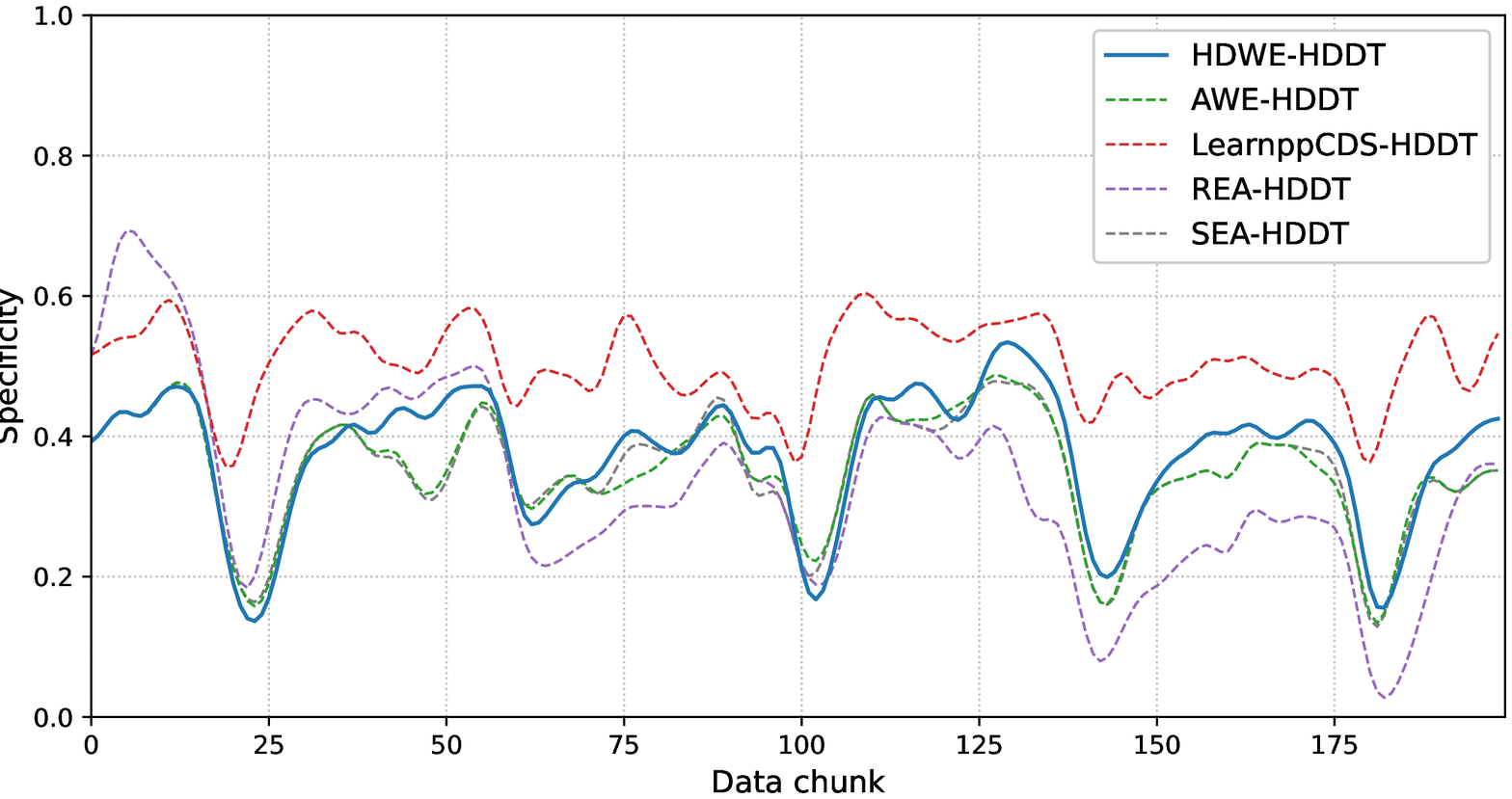}
\caption{Specificity} \label{fig:experiment2bsudB}
\end{subfigure}
\caption{Performance scores for state-of-the-art methods with base classifier \textsc{hddt} -- the generated data stream with the sudden concept drift and the stationary imbalance} \label{fig:experiment2bsud}
\end{figure}

\begin{figure}[!ht]
\centering
\begin{subfigure}{0.8\textwidth}
\includegraphics[width=\linewidth]{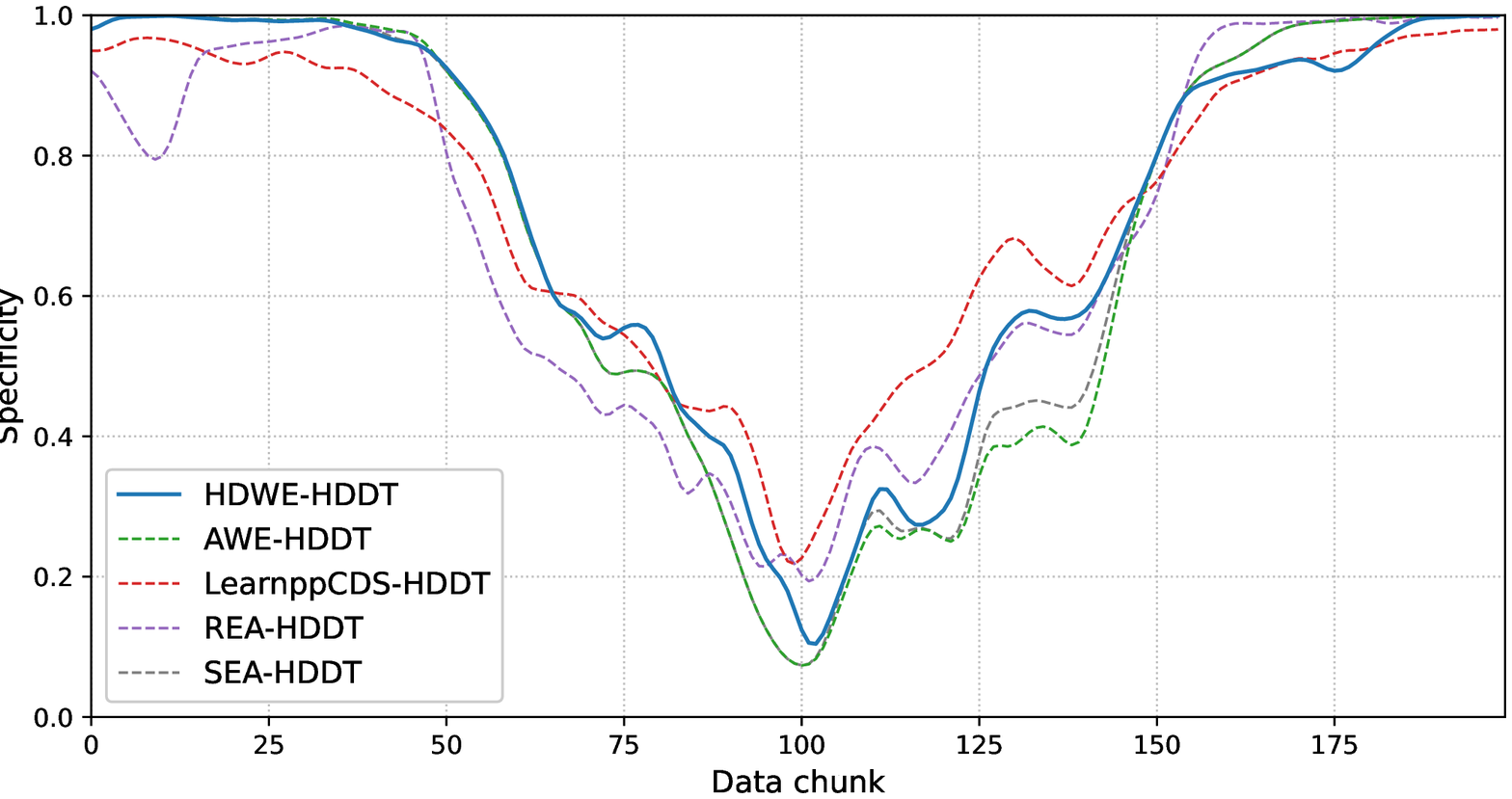}
\caption{Recall} \label{fig:experiment2bincA}
\end{subfigure}
\begin{subfigure}{0.8\textwidth}
\includegraphics[width=\linewidth]{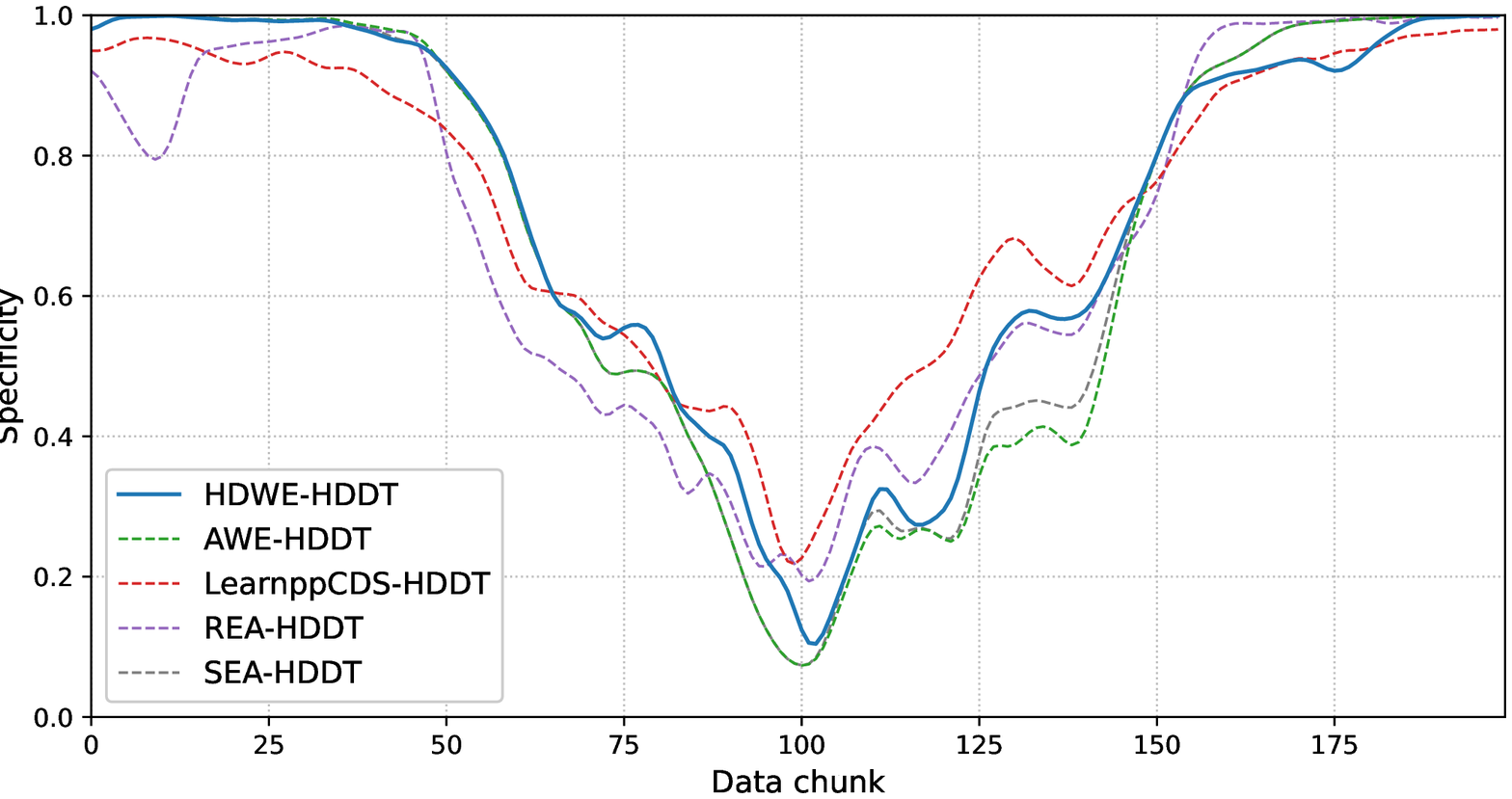}
\caption{Specificity} \label{fig:experiment2bincB}
\end{subfigure}
\caption{Performance scores for state-of-the-art methods with base classifier \textsc{hddt} -- the generated data stream with the incremental concept drift and the dynamically imbalance} \label{fig:experiment2binc}
\end{figure}

For the \textsc{hddt} base classifier, similar analyzes of the \textsc{hdwe} and selected state-of-the-art methods were performed. For two examples of the data stream and two metrics, Figure \ref{fig:experiment2bsud} and \ref{fig:experiment2binc} show performance. Data streams have the imbalance ratio 10\%. Figures \ref{fig:experiment2bsudA} and \ref{fig:experiment2bsudB} show sudden \textit{concept drifts} and the stationary imbalance. Figures \ref{fig:experiment2bincA} and \ref{fig:experiment2bincB} show incremental \textit{concept drifts} and the dynamical imbalance. It can be observed, that in each figure, the \textsc{hdwe} method slightly outperforms \textsc{awe}.

\begin{figure}[!ht]
\centering
\begin{subfigure}{0.49\textwidth}
\includegraphics[width=\linewidth]{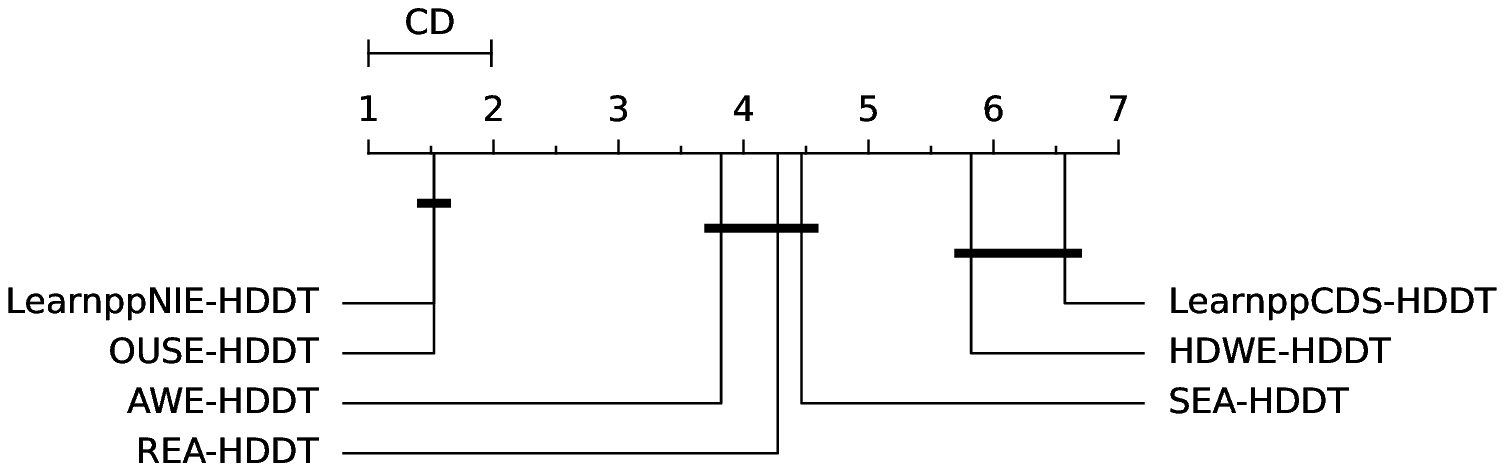}
\caption{Balanced Accuracy} \label{fig:ex2b_NemenyiA}
\end{subfigure}
\begin{subfigure}{0.49\textwidth}
\includegraphics[width=\linewidth]{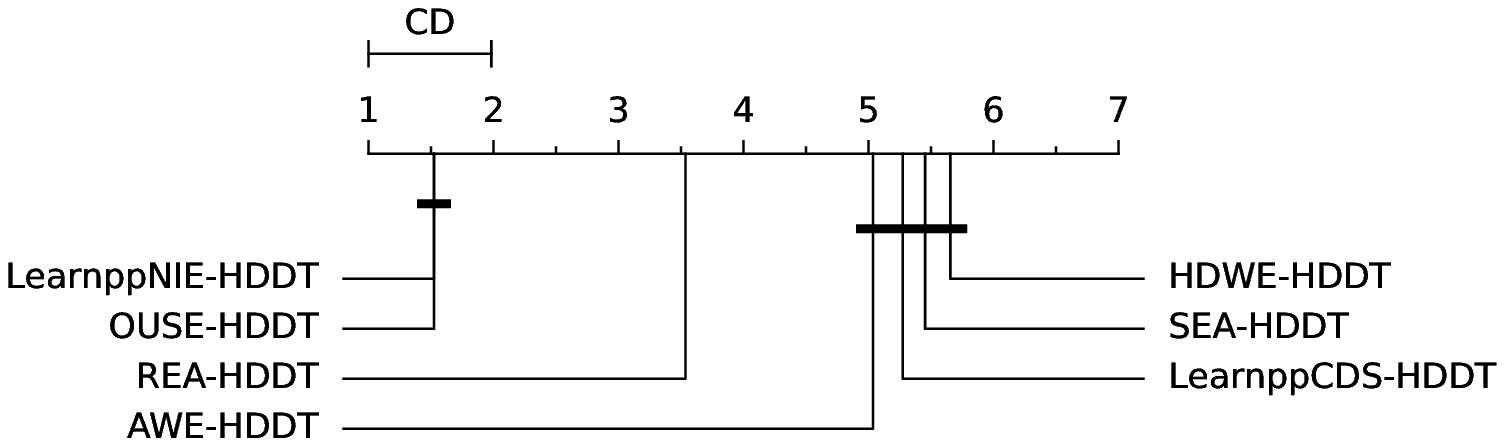}
\caption{$F_1$ score} \label{fig:ex2b_NemenyiB}
\end{subfigure}
\begin{subfigure}{0.49\textwidth}
\includegraphics[width=\linewidth]{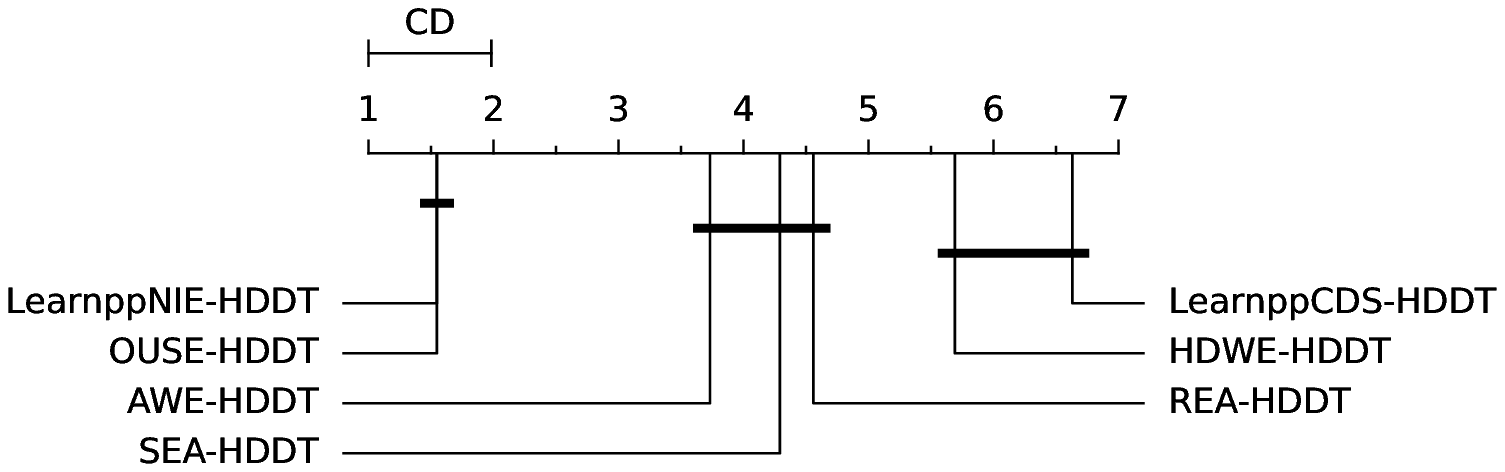}
\caption{G--mean} \label{fig:ex2b_NemenyiC}
\end{subfigure} 
\begin{subfigure}{0.49\textwidth}
\includegraphics[width=\linewidth]{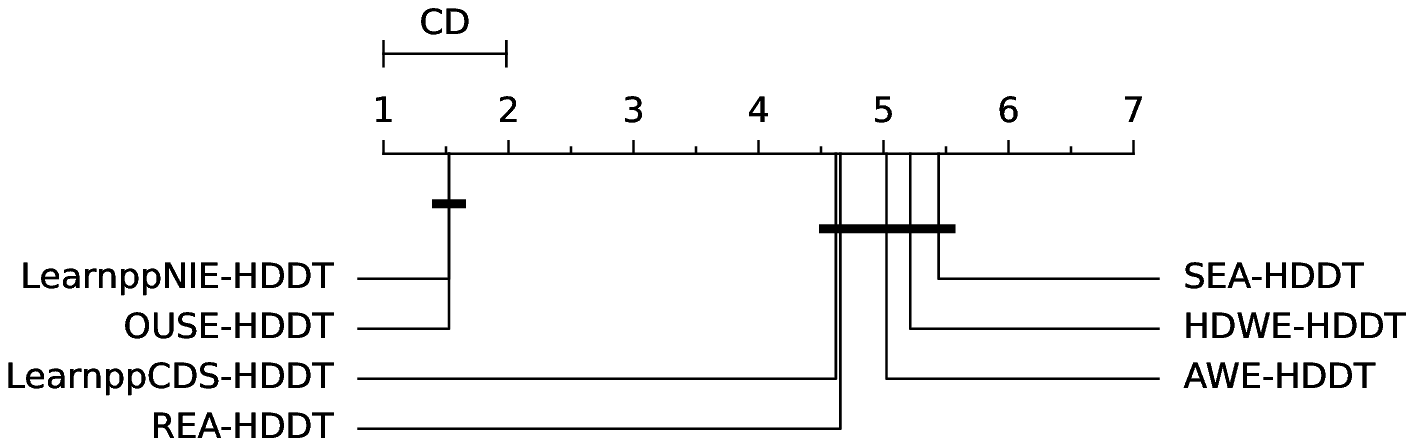}
\caption{Precision} \label{fig:ex2b_NemenyiD}
\end{subfigure}
\begin{subfigure}{0.49\textwidth}
\includegraphics[width=\linewidth]{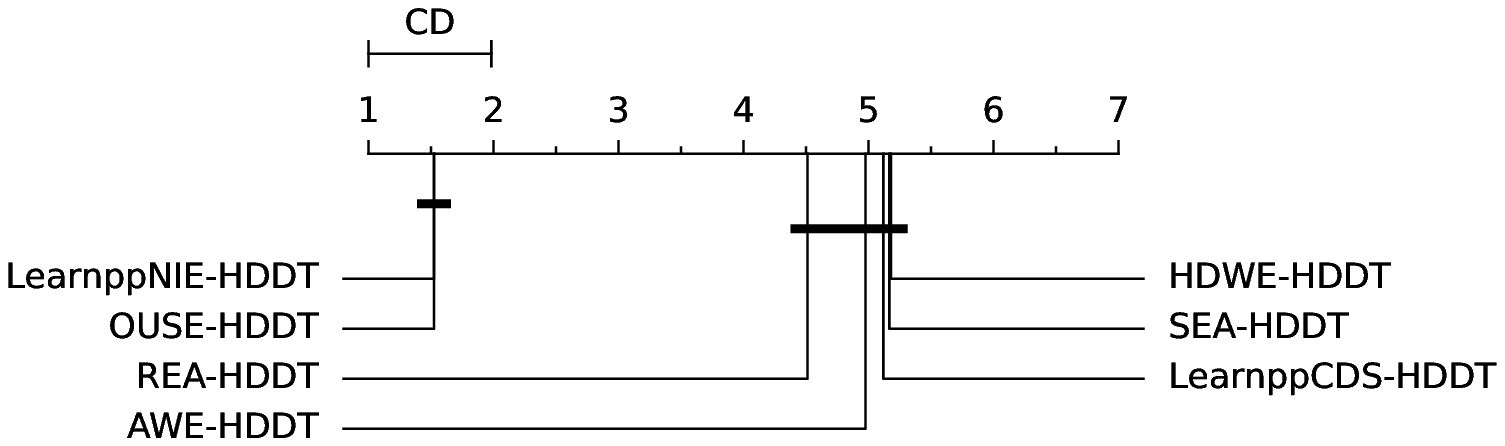}
\caption{Recall} \label{fig:ex2b_NemenyiE}
\end{subfigure}
\begin{subfigure}{0.49\textwidth}
\includegraphics[width=\linewidth]{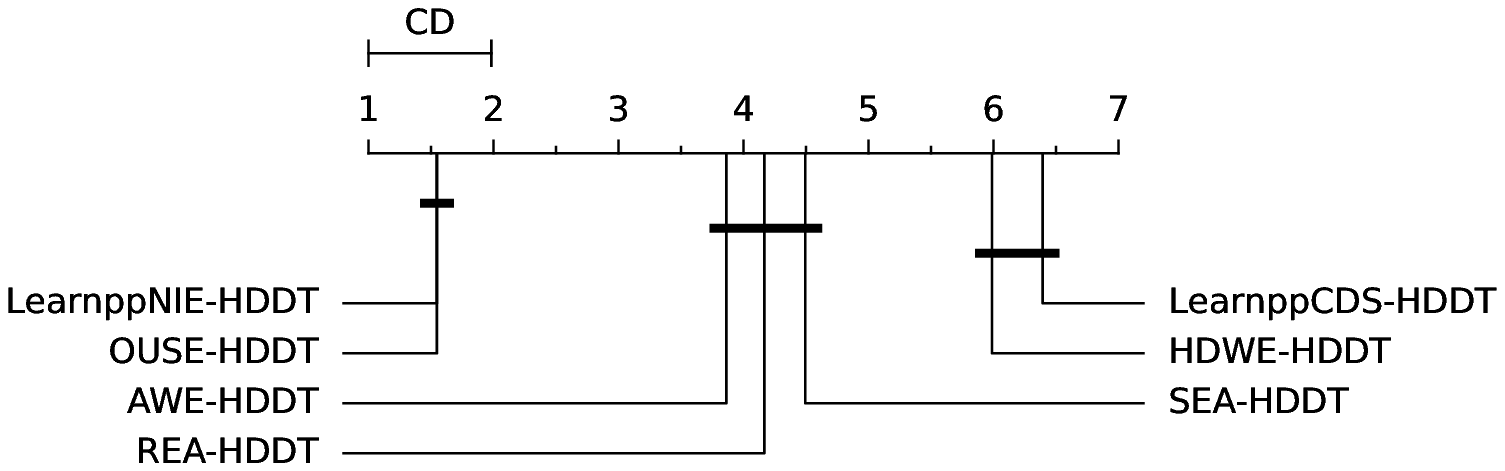}
\caption{Specificity} \label{fig:ex2b_NemenyiF}
\end{subfigure}
\caption{Diagrams of critical difference for Nemenyi test for state-of-the-art methods with the base classifier \textsc{hddt}}
\label{fig:ex2b-1_Nemenyi}
\end{figure}

Based on the average values of the stream metrics, the average ranks were calculated. Then, the \textit{Friedman test} is carried out. For p-value equal 0.05, the hypothesis H0 is rejected. The \textit{Nemenyi test} is shown in the Figure \ref{fig:ex2b-1_Nemenyi}. The confidence level $\alpha=0.05$ was used to calculate \textsc{cd}. The \textsc{hdwe} method has higher rank for \textit{$F_1$ score} and \textit{Recall}, while Learn++.\textsc{cds} is better for \textit{Balanced Accuracy}, \textit{G--mean}, \textit{Specificity}. For \textit{Precision}, the \textsc{sea} method has the biggest value. Methods \textsc{hdwe} and Learn++.\textsc{cds} are not different from each other. They are the best methods among others.

\begin{figure}[!ht]
\centering
\includegraphics[width=0.9\linewidth]{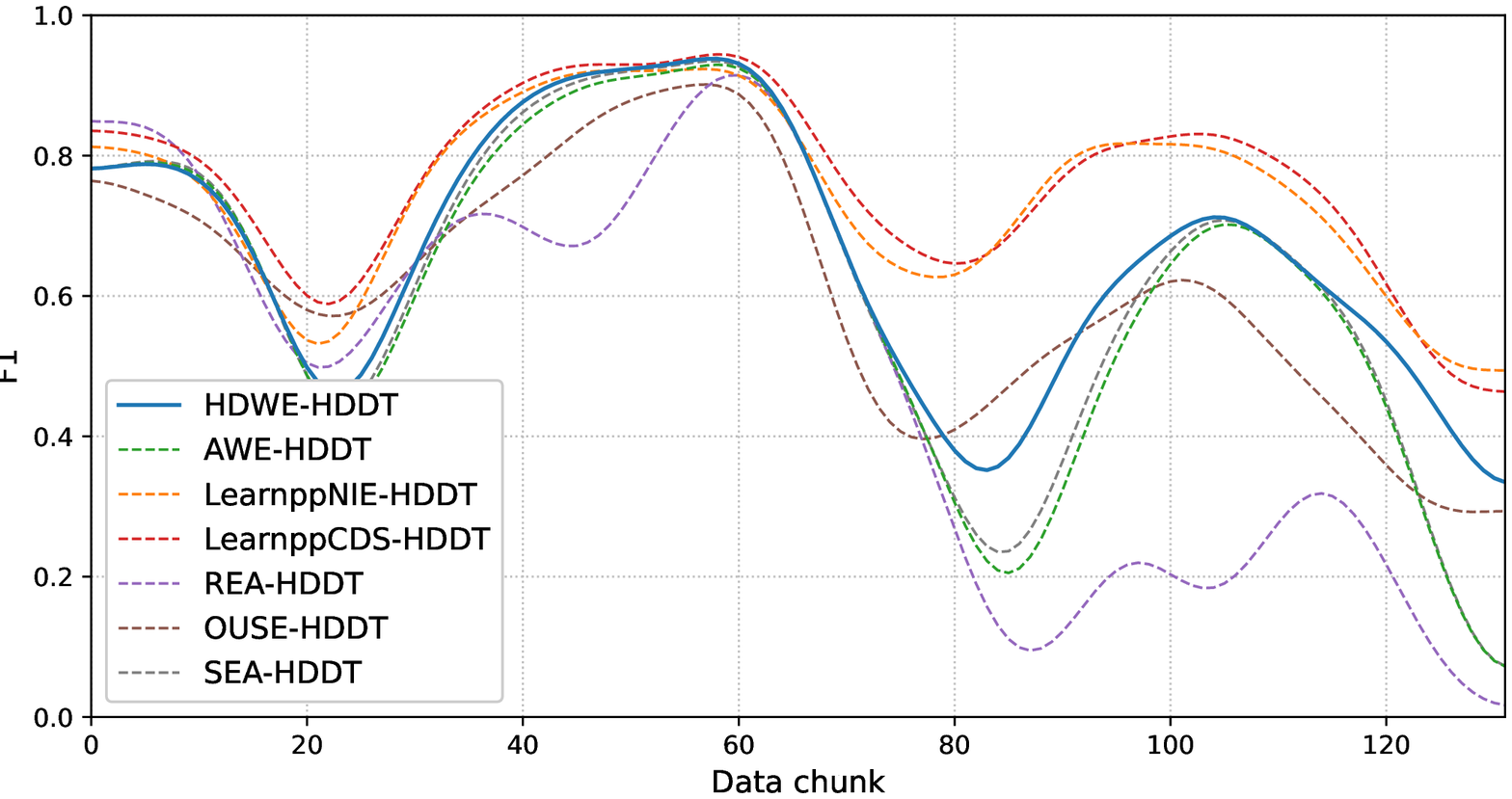}
\caption{$F_1$ score for state-of-the-art methods with base classifier \textsc{hddt} -- real data stream (\textit{covtype})} \label{fig:ex4_HDDT_covtype}
\end{figure}

\begin{figure}[!ht]
\centering
\includegraphics[width=0.9\linewidth]{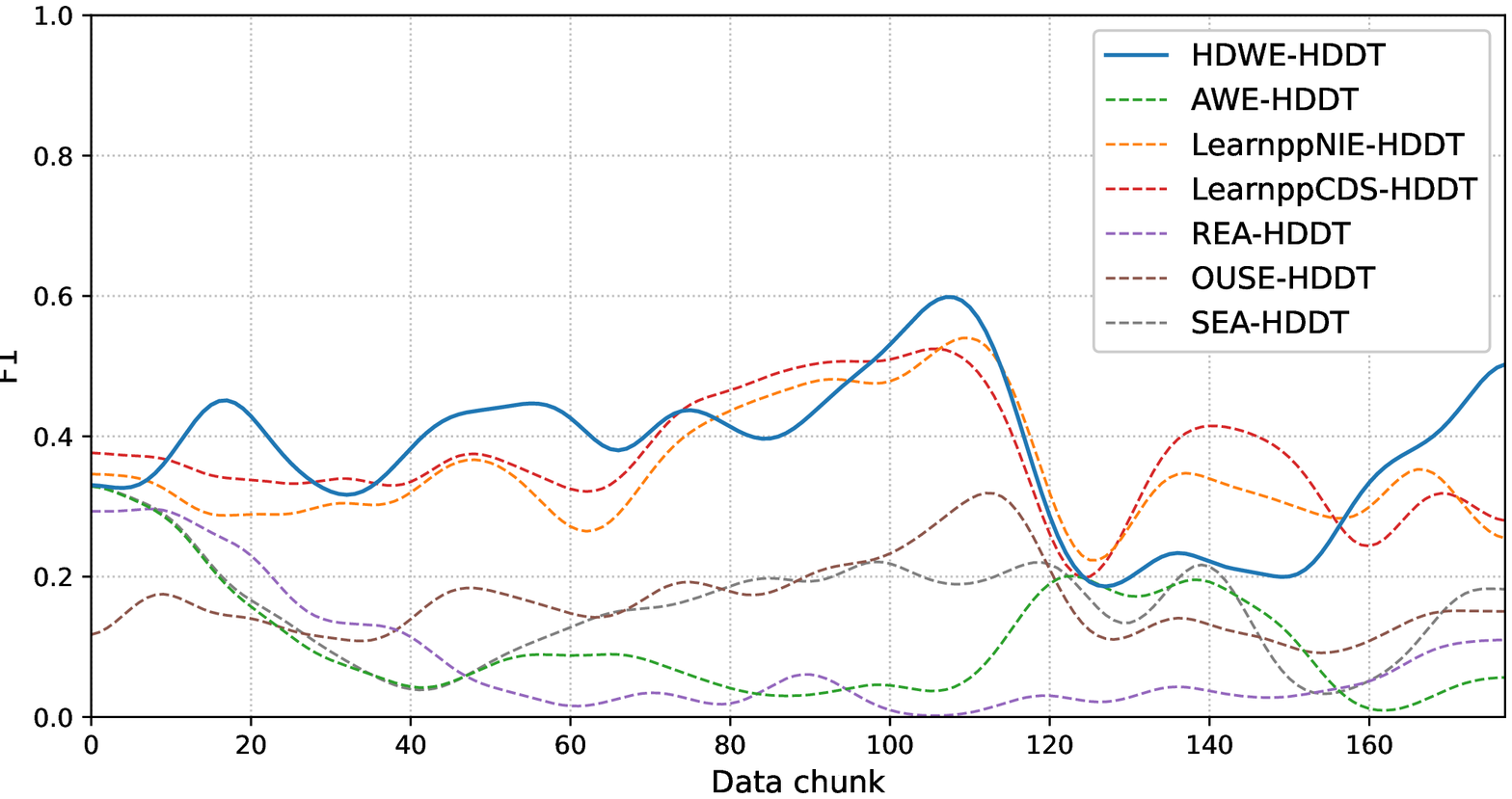}
\caption{$F_1$ score for state-of-the-art methods with base classifier \textsc{hddt} -- real data stream (\textit{poker})} \label{fig:ex4_HDDT_poker}
\end{figure}

\begin{figure}[!ht]
\centering
\includegraphics[width=0.9\linewidth]{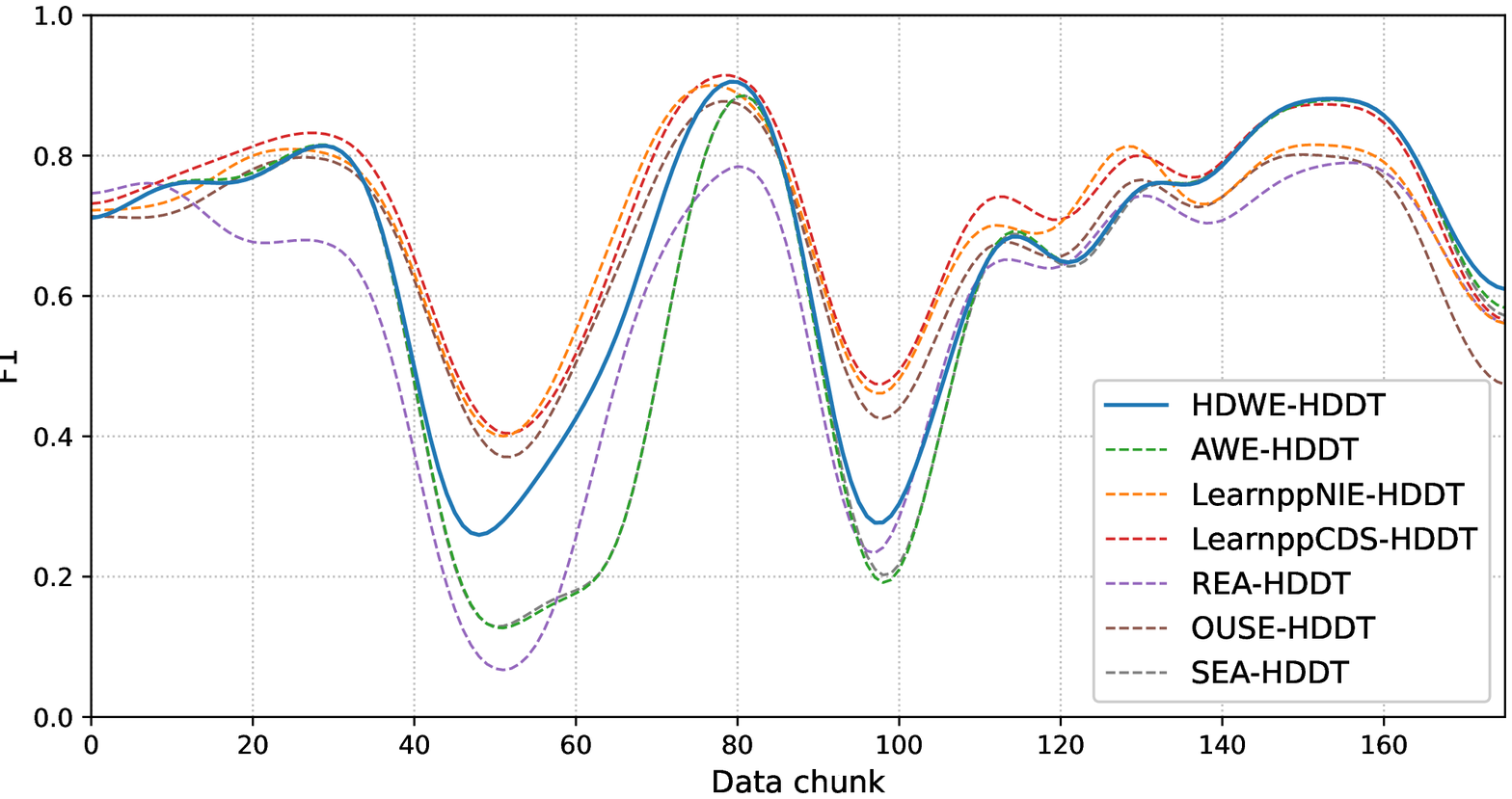}
\caption{$F_1$ score for state-of-the-art methods with base classifier \textsc{hddt} -- real data stream (\textit{insects})} \label{fig:ex4_HDDT_insects}
\end{figure}

\begin{figure}[!ht]
\centering
\begin{subfigure}{0.48\textwidth}
\includegraphics[width=\linewidth]{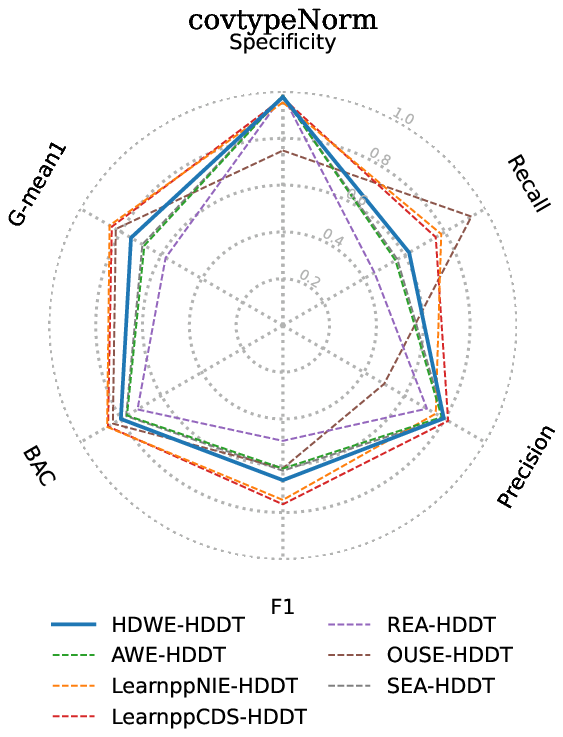}
\caption{\textit{\textit{Covtype}} data set} \label{fig:ex4_radar_HDDT_A}
\end{subfigure} \hfill
\begin{subfigure}{0.48\textwidth}
\includegraphics[width=\linewidth]{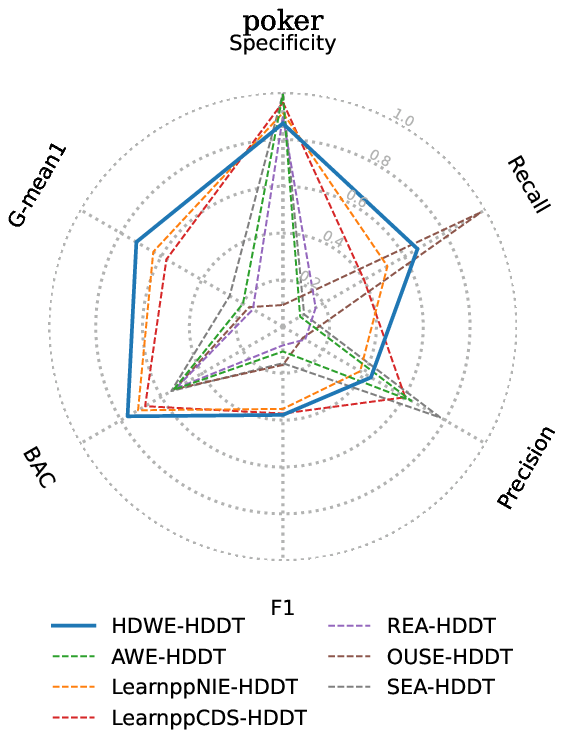}
\caption{\textit{Poker} data set} \label{fig:ex4_radar_HDDT_B}
\end{subfigure} \hfill
\begin{subfigure}{0.48\textwidth}
\includegraphics[width=\linewidth]{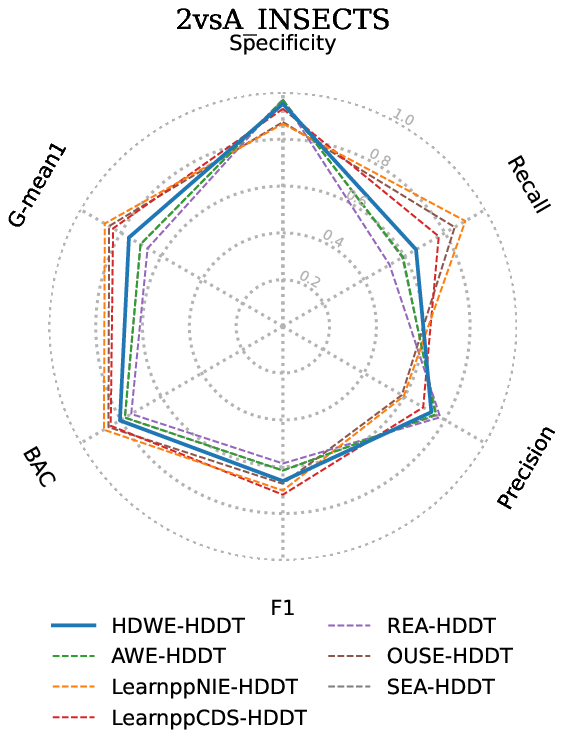}
\caption{\textit{Insects} data set} \label{fig:ex4_radar_HDDT_ii}
\end{subfigure}
\caption{Performance scores for state-of-the-art methods with base classifier \textsc{hddt} -- real data sets} \label{fig:ex4_radar_HDDT}
\end{figure}

Performance on real data sets \textit{covtype}, \textit{poker} and \textit{insects} is shown in Figures \ref{fig:ex4_HDDT_covtype}, \ref{fig:ex4_HDDT_poker} and \ref{fig:ex4_HDDT_insects} respectively. \textsc{hdwe} achieves the highest value of $F_1$ score while other methods have values from almost 0 to 0.4. Figure \ref{fig:ex4_radar_HDDT} shows performance based on all metrics. The \textsc{hdwe} method is as good as other in the classification of the \textit{covtype} and \textit{insects} data sets. Otherwise, it outperforms in the \textit{poker} data set for $F_1$ score, \textit{Balanced accuracy} and \textit{G--mean}.

\subsubsection{Imbalance comparison}

This section shows how the type of the imbalance affects the quality of the classification.

\begin{figure}[!ht]
\centering
\begin{subfigure}{0.8\textwidth}
\includegraphics[width=\linewidth]{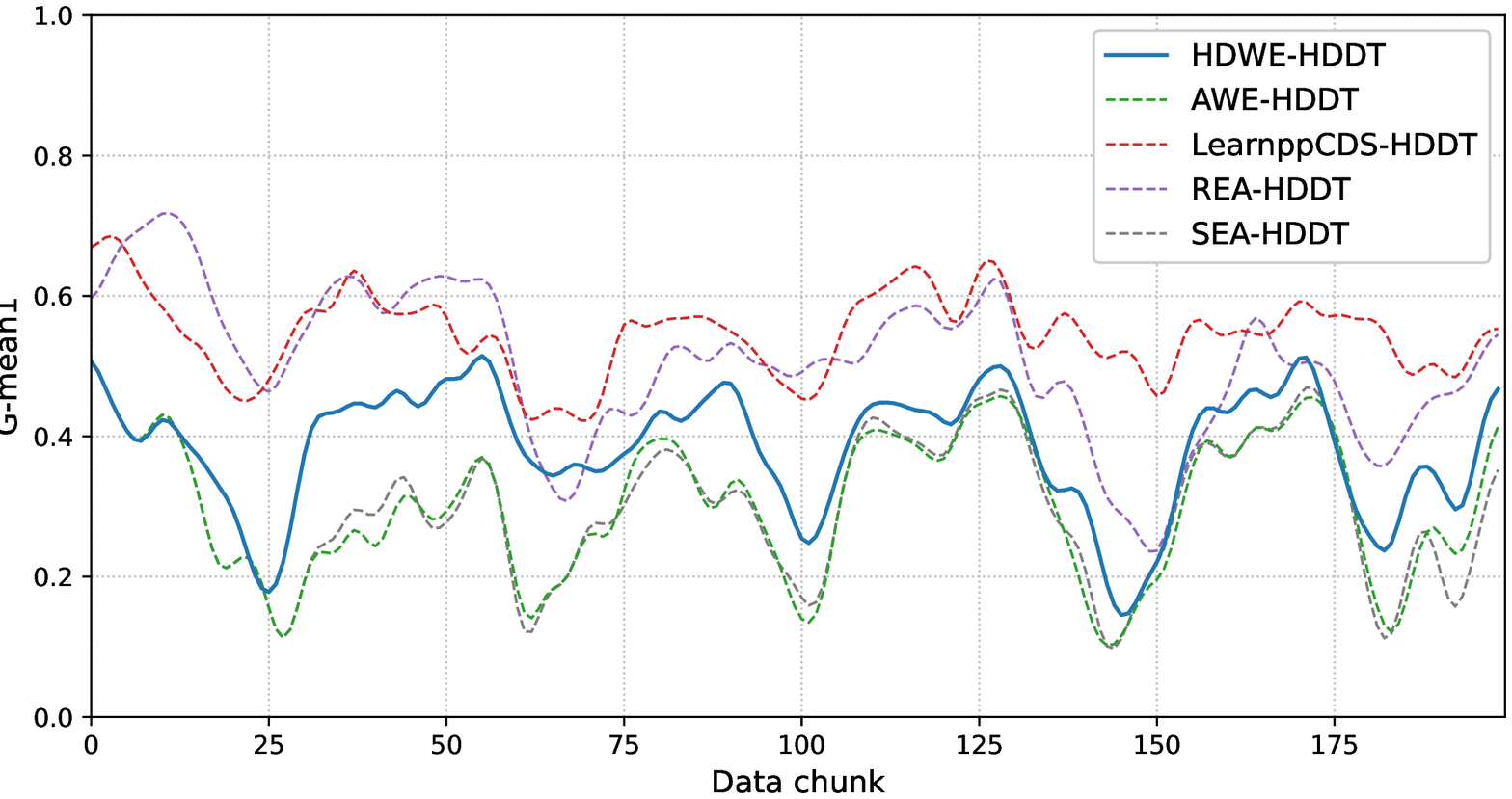}
\caption{Stationary imbalance} \label{fig:ex3imbincs}
\end{subfigure}
\begin{subfigure}{0.8\textwidth}
\includegraphics[width=\linewidth]{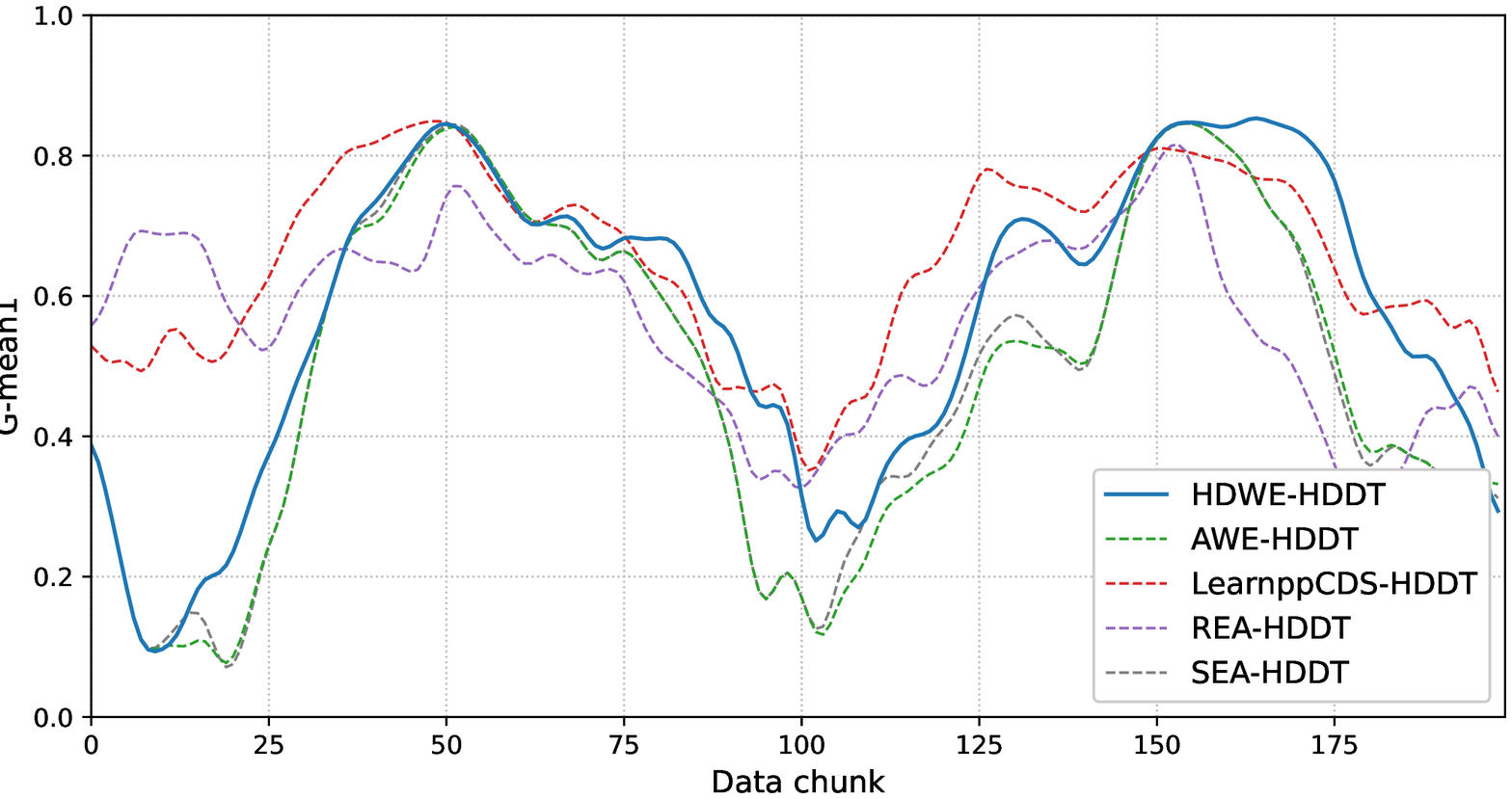}
\caption{Dynamically imbalance} \label{fig:ex3imbincd}
\end{subfigure}
\caption{G--mean score for the generated data stream with incremental concept drifts} \label{fig:ex3imbinc}
\end{figure}

An important point of this work is paying attention to the dynamic and stationary imbalance. Figure \ref{fig:ex3imbinc} shows \textit{G--mean} measure for the same stream with incremental \textit{concept drifts} and 5\% imbalance. Figure \ref{fig:ex3imbincs} clearly shows 5 drifts. It is a stream without changing the imbalance level. However, Figure \ref{fig:ex3imbincd} reflects the imbalance. The first class is 5\% and the second class is 95\%. In the 50th chunk, i.e., in 1/4 of the entire stream, classes' distribution is evened out. In the middle of the stream, the classes change, i.e., the first class is the majority, the second -- the minority, and the classification's quality decreases. Then, in 3/4 of the stream, the two classes are balanced again, and finally, the stream returns to the initial imbalanced state.

\begin{figure}[!ht]
\centering
\begin{subfigure}{0.8\textwidth}
\includegraphics[width=\linewidth]{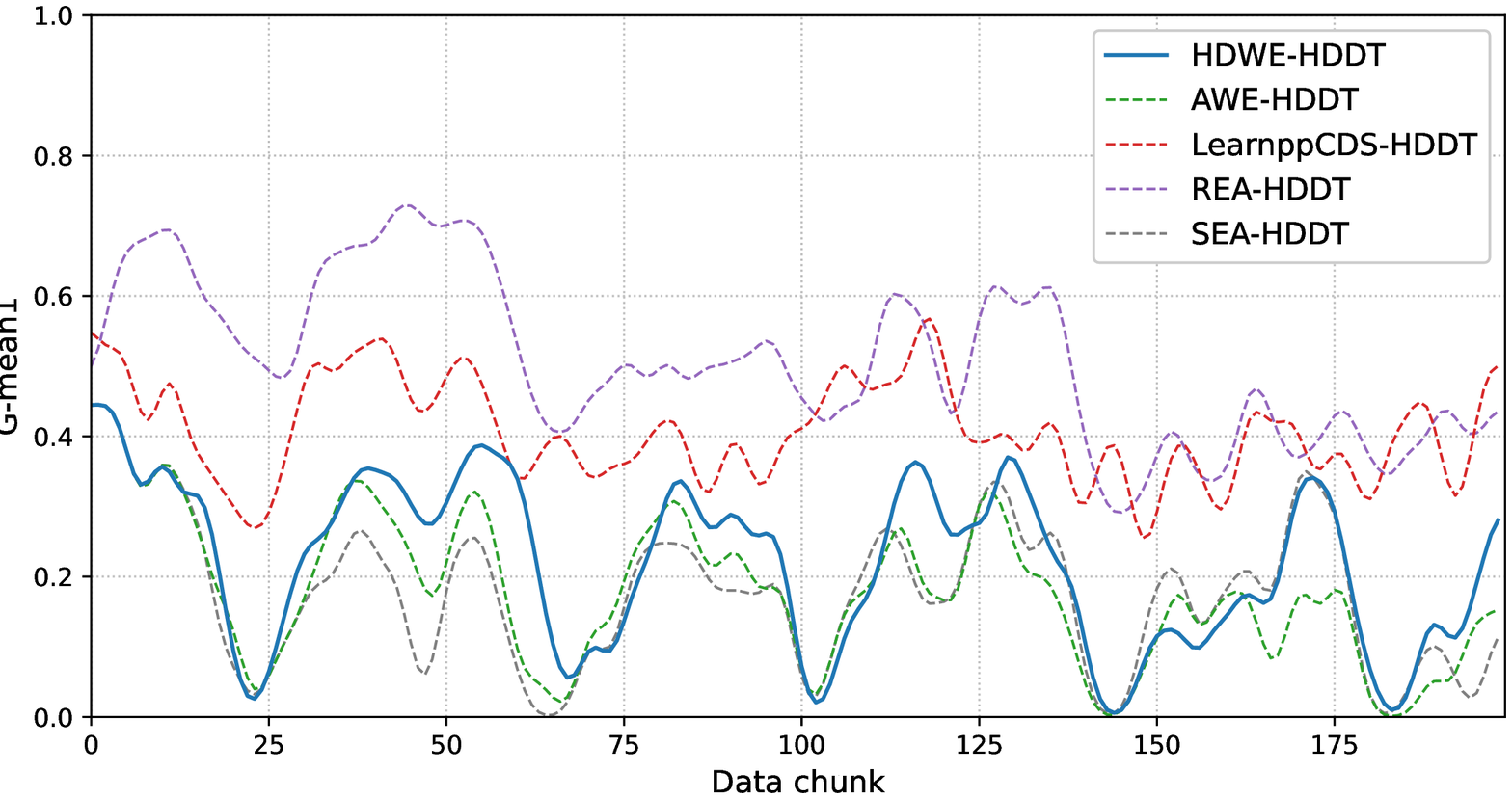}
\caption{Stationary imbalance} \label{fig:ex3imbsuds}
\end{subfigure}
\begin{subfigure}{0.8\textwidth}
\includegraphics[width=\linewidth]{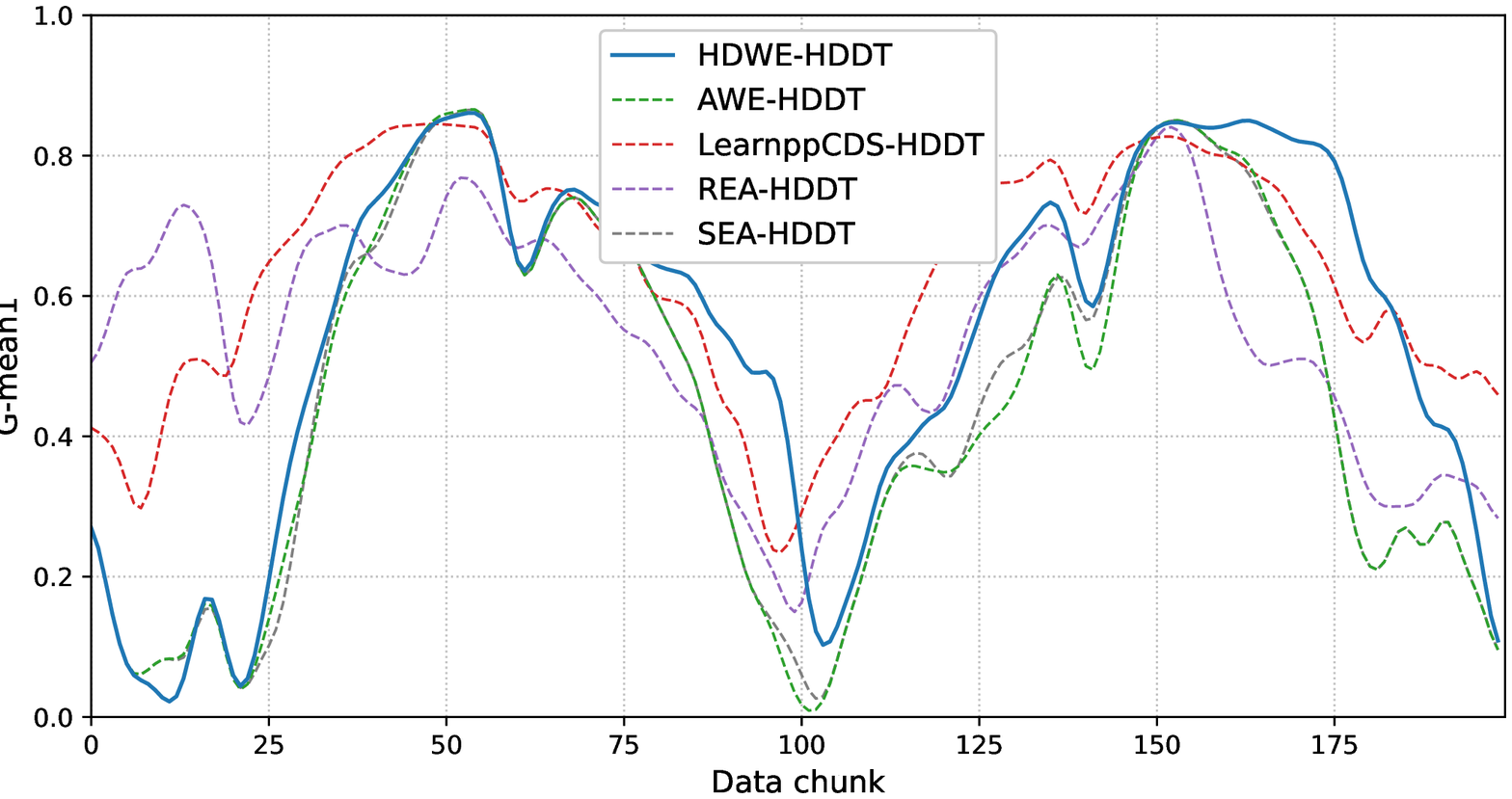}
\caption{Dynamically imbalance} \label{fig:ex3imbsudd}
\end{subfigure}
\caption{G--mean score for the generated data stream with sudden concept drifts} \label{fig:ex3imbsud}
\end{figure}

As before, Figure \ref{fig:ex3imbsud} shows \textit{G--mean} score, but this time for sudden \textit{concept drifts} and 3\% imbalance. In the case of dynamically imbalanced data stream (Figure \ref{fig:ex3imbsudd}), all algorithms work at a similar level.

\subsection{Lessons learned}

The \textit{Hellinger Distance Weighted Ensemble} is a typical chunk-based approach \cite{Krawczyk:2016}. The weight of the candidate in the ensemble depends on the \textit{Hellinger Distance}. Thanks to that, \textsc{hdwe} prefers the minority class and removes bias for the majority class because, in the case of imbalanced data, the cost of the minority class misclassification is higher. It is the algorithm-level method, so it does not use oversampling nor undersampling. Based on the experiments, the answers to research questions formulated in the beginning are as follows:

\begin{itemize}
    \item[RQ1:] \textit{How does the \textsc{hdwe} method work with different base classifiers?} \\
    The first experiment shows that the quality of the classification depends on the type of the base classifier. If high performance is expected for both the minority and majority classes, then \textsc{svc} or \textsc{mlp} should be chosen. 
    
    \item[RQ2:] \textit{Does the predictive performance of the \textsc{hdwe} outperform selected state-of-the-art methods?} \\
    It works as well as other selected state-of-the-art methods with non-stationary and imbalanced data streams. By comparing methods and averaging their results for all generated streams, i.e., for sudden and incremental \textit{concept drifts}, the stationary and dynamically imbalanced data stream, \textsc{hdwe} is statistically quite good compared to selected state-of-the-art methods. It shows high quality in classifying the minority class. This is probably because the \textit{Hellinger Distance} is used, which is insensitive to the imbalance. \textsc{hdwe} can also be used for the real data streams. Experiments showed that \textsc{hdwe} classified with comparable quality to other methods and slightly outperforms them, especially for the complex, difficult data.
    
    \item[RQ3:] \textit{How flexible is the \textsc{hdwe} in the non-stationary and dynamically imbalanced data sets?}\\
    The \textsc{hdwe} method can be used for dynamically imbalanced data because it achieves similar results to other selected methods. However, it is not completely resistant to changes in the prior probabilities of both \textit{concept drifts} and the variable imbalance.
    
\end{itemize}

\section{Conclusions}
\label{sec:conclusions}

This work proposed the \textit{Hellinger Distance Weighted Ensemble} (\textsc{hdwe}) method for batch learning and the binary classification data stream with occurring \textit{concept drifts} and the imbalance among classes. It is the classifier ensemble in which base classifiers are selected based on the value of the \textit{Hellinger Distance} calculated using the \textit{True Positive Rate} and the \textit{False Positive Rate}. When the ensemble's size is bigger than the number given initially, the worst model is removed. 
The computer experiments confirmed the satisfactory quality of the proposed algorithm in comparison to the state-of-art methods.

Future research could consider:

\begin{itemize}
    \item To examine the relationship between \textit{concept drifts} and the classification quality, and to use a drift detector to investigate a quality improvement time.
    \item Testing how the size of the ensemble affects metrics such as \textit{Balanced Accuracy}, \textit{$F_1$ score}, \textit{G--mean}, \textit{Precision}, \textit{Recall}, \textit{Specificity}.
    \item Inclusion advanced Neural Network as the base classifier for ensemble methods to improve the classification quality.
    \item Extension of the \textsc{hdwe} algorithm to multi-class problems and embed it into hybrid architectures with data preprocessing algorithms.
\end{itemize}


\bibliographystyle{elsarticle-num} 
\bibliography{bibliography}

\end{document}